\documentclass[sigconf]{acmart}

\renewcommand\footnotetextcopyrightpermission[1]{} 
\usepackage{graphicx,mathtools}
\usepackage{amsbsy}
\usepackage{bm}
\usepackage{bbm}
\usepackage{dsfont}
\usepackage{amsfonts}

\newcommand{\bftheta}{\pmb{\theta}}

\newcommand{\ind}[1]{\mathds{1}\left[#1\right]}
\newcommand{\calA}{\mathcal{A}}
\newcommand{\calU}{\mathcal{U}}

\newcommand{\calE}{\mathcal{E}}
\newcommand{\calG}{\mathcal{G}}
\newcommand{\bfy}{\mathbf{Y}}
\newtheorem{prop}{Property}
\DeclareMathOperator{\agg}{Aggregate}

\usepackage{url}

\begin{document}
%
\title{DeepGroup: Representation Learning for Group Recommendation with Implicit Feedback}

\author{Sarina Sajadi Ghaemmaghami}
\affiliation{
 \institution{Ontario Tech University}
 \country{Ontario, Canada}
 }
\email{sarina.sajadighaemmaghami@ontariotechu.ca}

\author{Amirali Salehi-Abari}
\affiliation{
 \institution{Ontario Tech University}
 \country{Ontario, Canada}
 }
\email{abari@ontariotechu.ca}

\begin{abstract}
Group recommender systems facilitate group decision making for a set of individuals (e.g., a group of friends, a team, a corporation, etc.). Many of these systems, however, either assume that (i) user preferences can be elicited (or inferred) and then aggregated into group preferences or (ii) group preferences are partially observed/elicited. We focus on making recommendations for a new group of users whose preferences are unknown, but we are given the decisions/choices of other groups. By formulating this problem as \emph{group recommendation from group implicit feedback},  we focus on two of its practical instances: \emph{group decision prediction} and \emph{reverse social choice}. Given a set of groups and their observed decisions, \emph{group decision prediction} intends to predict the decision of a new group of users, whereas \emph{reverse social choice} aims to infer the preferences of those users involved in observed group decisions. These two problems are of interest to not only group recommendation, but also to personal privacy when the users intend to conceal their personal preferences but have participated in group decisions. To tackle these two problems, we propose and study DeepGroup---a deep learning approach for group recommendation with group implicit data. We empirically assess the predictive power of DeepGroup on various real-world datasets, group conditions (e.g., homophily or heterophily), and group decision (or voting) rules. Our extensive experiments not only demonstrate the efficacy of DeepGroup, but also shed light on the privacy-leakage concerns of some decision making processes.

\end{abstract}




\maketitle

\pagestyle{plain}

\section{Introduction}
\label{section:intro}
Group decision problems are prevalent, ranging from high-stake decisions (e.g., elections, the board of directors' decisions, etc.) to casual decisions (e.g., deciding on a restaurant, movie, or vacation package for a group of friends). Nowadays, making a group decision in both online social platforms and face-to-face settings is challenging due to the overwhelming availability of options, and the complex and conflicting preferences of group members. In such scenarios, group recommender systems play an integral role in facilitating group decision making by recommending a set of items (or a single item) to a group of people such that the recommendation is satisfactory for all the members. The applications of group recommender systems are diverse, such as tourism \cite{mccarthy2006cats}, music \cite{crossen2002flytrap}, crowdfunding \cite{rakesh2016probabilistic}, news/web pages \cite{pizzutilo2005group}, TV programs \cite{yu2006tv}, and movies \cite{o2001polylens}. 

Group recommendation methods (e.g., \cite{BMR2010group,Amer-Yahia:2009,SYMM2011,GXLBH2010,POSN2015,XiaoRecsys2017,cao2018attentive}) usually encompass two key integrated components: \emph{preference assessment} and \emph{preference aggregation}. The \emph{preference assessment} component focuses on understanding group members’ preferences. Two common approaches for preference assessment are (a) \emph{preference elicitation} (e.g., \cite{kalech-elicitation:2011,Xia-Conitzer:aaai08,Lu-Boutilier_elicitation:ijcai11}) via asking relevant queries for revealing user preferences, and (b) \emph{preference learning} from historical user data represented either in the form of rankings (e.g.,\cite{hughes2015computing,POSN2015,doucette2015conventional}) or user-item interaction data (e.g.,\cite{Masthoff2004,BMR2010group,Amer-Yahia:2009,SYMM2011,GXLBH2010}). The \emph{preference aggregation} component aggregates individuals’ inferred (or elicited) preferences into group preferences (or decisions). These aggregation methods are usually well-studied social choice functions (or group consensus functions) \cite{compsoc16,felfernig2018group} or learned 
by deep learning (e.g.,\cite{cao2018attentive,yin2019social,huang2020efficient,hu2014}). In this paper, we deviate from this common approach of preference assessment and aggregation for group recommendation.  

We focus on a group recommendation problem in situations that the group members' personal preferences cannot be assessed using typical means (e.g., preference elicitation or preference learning) due to unavailability or confidentiality of preferences. However, we assume the presence of a certain type of \emph{implicit feedback} for some group members in the form of which other groups they belong to and those groups' decisions. The applications for this problem are prevalent, e.g., group recommendations for restaurants and vacation packages given that we can observe (through online social platforms) restaurants and places that some group members have attended with their friends or family members. We focus on two special instances of this problem: \emph{group decision prediction}, which intends to predict the decision of a new group of users, and \emph{reverse social choice}, which aims to infer the preferences of a user involved in observed group decisions.\footnote{The latter special case is derived when the new group is a singleton set.}  In addition to group recommendation, these two special cases are of high importance for assessing privacy leakage. Imagine those users who intend to conceal their personal preferences on a sensitive issue (e.g., promotion, social injustice issues, etc.), but have participated in group decisions on these topics with publicly known decisions.  

\vspace{1mm}
\noindent \textbf{Contribution.} To address the group recommendation problem with group implicit feedback, we propose \emph{DeepGroup}---a deep neural network for learning group representations and decision making. We conducted extensive experiments to evaluate the effectiveness of DeepGroup for both group decision prediction and reverse social choice. Our findings confirm the superiority of DeepGroup over a set of benchmarks for both problems over various datasets. In our experiments, we also study how different group decision rules (or group decision-making processes) might affect the performance of DeepGroup. Our findings show that DeepGroup excels (compared to benchmarks) regardless of the choice of group decision rule and even perform reliably well when different unknown decision rules are used in observed groups. In the reverse social choice task, DeepGroup performance was more prominent for plurality voting. This is an interesting observation regarding privacy. Despite requiring the least personal preference data (i.e., only top choice) for decision making, plurality has the highest privacy leakage. 

\section{Related Work}
\label{section:related}

We review the related work on group recommender systems, computational social choice, preference learning, and deep learning recommender systems. 

\vskip 1.5mm
\noindent \textbf{Group Recommender Systems.}
Group recommendation is an emerging area in recommender systems \cite{felfernig2018group,dara2020survey}. The group recommendation methods broadly fall into three categories: (i)
\emph{Artificial profile} methods \cite{mccarthy1998musicfx,kim2010group} create a joint virtual user profile for a group of users to keep track of their joint revealed/elicited preferences;
(ii) \emph{Profile-merging} methods \cite{yu2006tv,BF2010Rec} form a group profile by merging its members' user-item interactions, then recommendation will be made based on a group profile; and (iii) \emph{Recommendation/preference
aggregation} methods \cite{Masthoff2004,BMR2010group,Amer-Yahia:2009,SYMM2011,GXLBH2010,POSN2015,XiaoRecsys2017} aggregate the recommendations (or
inferred preferences) of group members into a group recommendation by using various \emph{social choice functions} (also referred to as \emph{group consensus functions}). There has been a growing interest in applying  deep learning for group recommendations \cite{hu2014,cao2018attentive,yin2019social,huang2020efficient,GroupIM2020}. GroupIM \cite{GroupIM2020} learns preference aggregators for ephemeral groups. Neural networks with attention mechanisms are also deployed for learning the aggregation strategies (i.e., group consensus functions) of group member's (predicted) preferences \cite{cao2018attentive,yin2019social,huang2020efficient}.

For a group recommendation, all these methods assume that user preferences/profiles of group members or group preferences are accessible. Our goal differentiates from this body of research. We make a group recommendation to a new group of users whose preferences (or profiles) were not observed, but their participation in group decision making of some other groups has been observed.

\vskip 1.5mm
\noindent \textbf{Social Choice and Preference Learning.}  
Social choice equips group recommender systems with a principled framework for aggregating individuals' preferences into group preferences (or decisions) \cite{compsoc16}. Many social choice schemes (such as voting rules) have been studied \cite{compsoc16}. The impact of social relationships among group members on group recommender systems has studied further. The problem of preference assessment in social choice settings is very predominant as a collection of user preferences must be elicited or learned to make a group decision. The research has focused on two approaches of \emph{preference elicitation} via asking relevant queries which result in revealing user preferences and \emph{preference learning} from historical user-item interaction data. The preference elicitation methods for social choice problems has been developed based on diverse approaches ranging from heuristic methods using the concept of \emph{possible and necessary winners} \cite{kalech-elicitation:2011,konczak-lang:preferences05,Xia-Conitzer:aaai08} to the notion of \emph{minimax regret} \cite{Lu-Boutilier_elicitation:ijcai11,Lu-Boutilier_multiwinner:ijcai13}. There is also growing interest in predictive models of preferences for learning user preferences in social choice settings (for example, see \cite{hughes2015computing,POSN2015,doucette2015conventional}).

Our work differs in several ways. Rather than eliciting or inferring missing user preferences for group decision making, we predict the group decision (or preferences) from some other groups' observed decisions. A special instance of our problem, reverse social choice, has the opposite goal of social choice by segregating user preferences from group decisions.  

\vskip 1.5mm
\noindent \textbf{Recommender Systems and Neural Networks.} 
Deep learning has recently shown enormous potential in improving recommender systems by capturing enriched user and item representations \cite{ZYST2019}. Different types of neural networks are successfully applied for top-k recommendation such as feedforward networks \cite{wideDeep2016,he2017neural}, recurrent networks \cite{wu2017recurrent}, classical autoencoders \cite{ZWC-JCA-2019}, denoising autoencoders \cite{wu2016collaborative}, and variational autoencoders \cite{LS-2017-KDD,liang2018variational,askari2020joint}. However, these models can't directly be applied to our group recommendation problem with implicit feedback as their emphasis is on predicting user preferences from user-item interactions.  

\vskip 2mm
\noindent \textbf{Group Decision Making and Social Factors.}
There has been a growing interest in studying and modeling social factors or interaction patterns of group members for group recommendation tasks \cite{delic2017comprehensive,MG-UMUAI2006,QSTIST2013,empathetic2014,delic2018use}. It is shown that the group member relationship types affect their emotional conformity and contagion \cite{MG-UMUAI2006}.  The strength of social relationships has been studied for group collaborative filtering problems \cite{QSTIST2013}. Empathetic preferences, in which an individual's satisfaction is derived from both its own and groupmate's satisfaction, has studied in the group recommendation context \cite{empathetic2014}. A recent study suggests that groups with socially-close group members are usually happier with the recommended group decision than those groups with loosely-related members \cite{delic2018use}. A series of observational studies are conducted, mainly in the tourism domain, to investigate the relationships between group members' characteristics, conflict resolution attitudes, and the satisfaction of group members from group decision \cite{delic2017comprehensive,delic2016observing}. Our work is loosely related to this literature by going beyond simple aggregation methods. However, our work differentiates from this literature by assuming that the individuals' preferences, group dynamics, and social factors are not observed.  

 


\section{Group Recommendation from Group Implicit Feedback}
\label{section:problem}
Our goal is to make a recommendation to a group of users whose personal preferences are unknown to the system. However, these users might have participated in group decision making processes of some other groups with known/public decision outcomes. The applications of this problem are prevalent in our daily lives. For instance, recommendations of restaurants or vacation packages to a group, when we have observed restaurants or places which they have visited with some other friends or family members.

We consider a set of $n$ users $\calU = \{1,\ldots, n\}$ and a set of $m$ alternatives (items) $\calA = \{a_1,\ldots, a_m\}$. We assume that we have observed $l$ groups of users $\calG = \{G_1,\ldots,G_l\}$ with $G_i \subseteq \calU$, and their corresponding group decisions (choices). The observed group decisions can be represented as the \textit{group-item interaction matrix} $\bfy=[y_{ij}] \in \{0,1\}^{l\times m}$, where $y_{ij}=1$ if $G_i \in \calG$ has decided $a_j \in \calA$ as their group decision.\footnote{One can extend our problem to the setting in which observed decisions/outcomes are in the form of group aggregated rankings rather than a consensus option.} In this setting, one can focus on the top-$k$ recommendation problem by suggesting the $k$ most preferred (or likely) items from $\calA$ to a new group of users $G \subseteq \calU$ where $G \notin \calG$.\footnote{The set of alternatives doesn't have to be the same for all group decisions. One can assume a universal set of alternatives with all possible alternatives. Regardless of which alternative has been available from the universal set for past group decisions, the alternative selected by a group signals the preferences of the group members.} While our defined problem covers a broad range of problems, of particular interest, are two special instances of it.

\vskip 2mm
\noindent \textbf{Group Decision Prediction.}  
Single-option group recommendation (i.e., when $k=1$)---sometimes referred to as \textit{group decision prediction}---not only applies to group recommendation but also can be used for predicting the most likely decision (or outcome) of a newly formed group $G$. Imagine a committee is asked to decide on a sensitive issue (e.g., promotion, social injustice issues, etc.) when various decisions are possible. The goal is to predict the final decision of the committee based on the involvement of the committee members in previous committees whose final decisions are public.     

\vskip 2mm
\noindent \textbf{Reverse Social Choice.} 
By letting the target group $G$ be a singleton set of a user $u \in \calU$, one can focus on a special instance of our problem, that we call \textit{reverse social choice.} As opposed to social choice functions that aggregate individuals' preferences to group decisions or group preferences, the reverse social choice intends to map group decisions to individuals' preferences. The solution to this problem not only helps us to enhance preference learning but also allow us to measure privacy leakage from publicly announced group decisions.  

Regardless of our interest in these two special instances of the group recommendation problem, our proposed solution is for the general problem and applicable to any instances. Our general approach is to predict the likelihood of the interaction of group $G$ with any item in $\calA$ (or preference of group $G$ over $\calA$), and then select a rank list of $k$ items with the highest prediction score for recommendation to the group $G$. Our learning task in this paper is to find the \textit{likelihood function} $f(G,a|\bftheta)$ that predicts the likelihood of group $G$'s interaction with any item $a \in \calA$. Here, $\bftheta$ denotes the model parameters and can be learned (or estimated) from the observed groups $\calG$ and group-item interaction matrix $\bfy$. We propose DeepGroup in Section \ref{section:deepGroup} as a powerful deep learning model for formulating and learning this likelihood function.   

\section{DeepGroup Model}
\label{section:deepGroup}
We propose DeepGroup neural network to address the group recommendation problem discussed in Section \ref{section:problem}. DeepGroup, by learning the likelihood function $f(G,a|\bftheta):2^\calU\times \calA \rightarrow [0,1]$ for $G \subseteq \calU$ and any $a \in \calA$, can be used to predict the likelihood of group $G_i$'s interaction with any item $a_j \in \calA$ by $\hat{y}_{ij} = f(G_i,a_j|\bftheta)$. We discuss the model architecture of DeepGroup, its key components (e.g., aggregator functions), and its learning process in this Section.  
\begin{figure}[tb]
\begin{center}
  \includegraphics[width=0.36\textwidth]{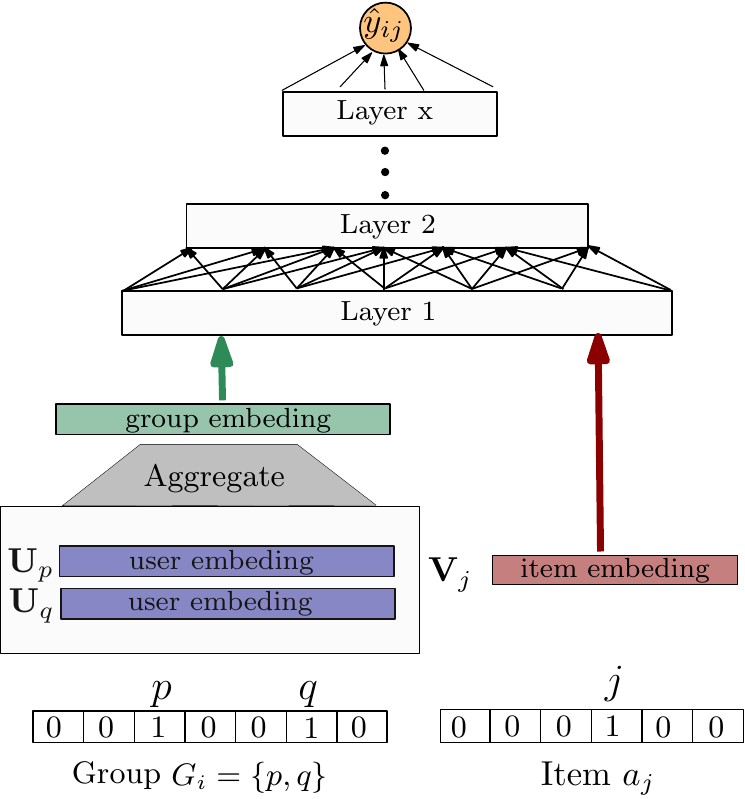}
 \vspace{-6pt}
  \caption{The architecture of the DeepGroup model.}
  \label{fig:deepGroupmodel}
\end{center}
\vspace{-18pt}
\end{figure}

\subsection{Model Architecture}
Figure \ref{fig:deepGroupmodel} depicts the architecture of DeepGroup. The DeepGroup model takes both group $G_i \subseteq \calU$ and item $a_j \in \calA$ as an input. The $G_i$ is represented as the $n$-row sparse vector $\mathbf{g} = [g_p]$ where $g_p=1$ if $p\in G_i$ otherwise $g_p=0$. The item $a_j$ can be represented by one-hot encoding in $m$-row vector $\mathbf{a}$. 

The DeepGroup considers real-valued latent representations (or embedding) for all users $u \in \calU$ and items $a \in \calA$. The latent representations of users and items are captured by $n\times d$ matrix $\mathbf{U}$ and  $m\times d'$ matrix $\mathbf{V}$ (resp.), where $d$ and $d'$ are the dimensions of user and item latent spaces (resp.). For the input group $\mathbf{g}$, DeepGroup retrieves all its users' latent representations 
 $\{\mathbf{U}_p | p \in \calU \text{ and } g_p=1\},$
where $\mathbf{U}_p$ denotes latent vector of user $p$ (i.e., the $p^{th}$ row in the matrix $\mathbf{U}$). Similarly, DeepGroup looks up the item embedding $\mathbf{V_j}$ for input item $a_j$. A key idea behind DeepGroup is the aggregation of a group $\mathbf{g}$'s users latent representations $\{\mathbf{U}_p | p \in \calU \text{ and } g_p=1\}$ into a single fixed-length vector $\mathbf{q}$:
\begin{equation}
    \mathbf{q} = \agg\left(\left\{\mathbf{U}_p | p \in \calU \text{ and } g_p=1\right\}\right).
\end{equation}
The $\agg(.)$ function takes any set of user latent representations and maps them into $\mathbf{q}$, which is the latent representation of the group $\mathbf{g}$. This group latent representation is expected to capture the consensus preference of group members. We discuss about different aggregator functions in Section \ref{sec:agg}.

The group latent representation $\mathbf{q}$ and the item embedding $\mathbf{V}_j$ are then concatenated and fed into a multilayer perceptron (MLP) neural network to predict $\hat{y}_{ij}$ (i.e., the likelihood that group $g$ decide on item $a_j$). The MLP consists of $X$ hidden layers formulated by
\begin{align*}
    \mathbf{h}^{(1)} &= f^{(1)}\left(\mathbf{W}^{(1)}\left(\mathbf{q} \mathbin\Vert \mathbf{V}_j\right) + \mathbf{b}^{(1)}\right)\\
    \mathbf{h}^{(i)} &= f^{(i)}\left(\mathbf{W}^{(i)}\mathbf{h}^{(i-1)} + \mathbf{b}^{(i)}\right)~~~~~~~~\text{for } i > 1,
\end{align*}
where $\mathbf{h}^{(i)}$, $\mathbf{W}^{(i)}$, $\mathbf{b}^{(i)}$, and $f^{(i)}(.)$ are the hidden unit vector, the linear transformation weight matrix, bias vector, and non-linear activation function for layer $i$, respectively. Here, $\mathbin\Vert$ is the concatenation operator. Finally, DeepGroup outputs 
\begin{equation}
    \hat{y}_{ij} = \sigma \left(\mathbf{w}_o\mathbf{h}^{(X)} + b_o\right),
\end{equation}
where $\sigma(.)$ is the sigmoid function for converging the linear transformation of the last hidden layer output $\mathbf{h}^{(X)}$ into a probability. Here, $\mathbf{w}_o$ and $b_o$ are the weight vector and bias parameter for the output layer.

\subsection{Aggregator Functions}
\label{sec:agg}
An integral part of DeepGroup is the aggregate function which maps any arbitrary set of user embeddings into group representation $\mathbf{q}$. A candidate function is required to satisfy at least two natural properties. 
\begin{prop}
A function $\agg$ acting on sets of user embeddings must be \textbf{permutation invariant} to the order of user embeddings in the set such that for any permutation $\pi$ and any user embedding set $\{\mathbf{U}_{i_1}, \cdots, \mathbf{U}_{i_j}\}$: $$\agg\left(\left\{\mathbf{U}_{i_1}, \cdots, \mathbf{U}_{i_j}\right\}\right) = \agg\left(\left\{\mathbf{U}_{\pi(i_1)}, \cdots, \mathbf{U}_{\pi(i_j)}\right\}\right).$$
\end{prop}

\begin{prop}
A function $\agg$ acting on sets of user embeddings must have a \textbf{fixed-length range} for any set of user embeddings: letting $\calE=\{\mathbf{U}_i | i\in \calU\}$ be the set of all users' embeddings, the function $\agg:2^{\calE}\rightarrow \mathbb{R}^k$ maps any subset of $\calE$ to a $k$-dimension real-valued vector. 
\end{prop}
Given these two properties, one can consider two classes of aggregate functions.

\vskip 2mm
\noindent \textbf{ELementwise Aggregators.} By deploying an elementwise operator (e.g., mean, max, min, median), an elementwise aggregator reduces a group $G$'s user embeddings $\{\mathbf{U}_p | p \in G\}$ into a group embedding $\mathbf{q}$. This class of aggregators generates the group embedding with the same dimensionality of user embedding. An important instance of this class is Mean aggregator which compute the $i^{th}$ element of the group embedding $\mathbf{q}$ by: 
\begin{equation}
    q_i  = mean(\{u_{pi} | p \in G\}),
    \label{eq:mean-agg}
\end{equation}
where $u_{pi}$ denotes the value of the $i^{th}$ dimension of user $p$'s embedding.   A variant of this Mean aggregator has been widely deployed in representation learning on graphs (for example, see \cite{hamilton2017inductive}) to aggregate features
from a node’s neighborhood. 

By replacing mean in Eq.~\ref{eq:mean-agg} to any other elementwise operators (e.g., min, max, median, etc.), one can derive other elementwise aggregator functions.    

\vskip 2mm
\noindent \textbf{Combined Aggregators.}
While an elementwise aggregator (e.g., Mean aggregator) can reduce a set of user embeddings into a single group embedding, it is possible that two distinct sets of user embeddings result in the same group embedding under a specific aggregator. 

To address such issues and make group representations more distinctive, one can combine multiple aggregators by concatenating their outputs. For example, we can define the mean-max-min aggregator for aggregating a set of user embeddings $\mathbf{U}_G=\{\mathbf{U}_p | p \in G\}$ by
\begin{equation}
    MMM(\mathbf{U}_G) = Mean(\mathbf{U}_G)\mathbin\Vert Max (\mathbf{U}_G) \mathbin\Vert Min(\mathbf{U}_G),
\end{equation}
where $\mathbin\Vert$ is the concatenation operator, and Mean, Max, and Min are elementwise aggregator functions. This combined aggregator has a fixed-length range with three times larger dimensionality of user embeddings. The mean-max-min aggregator has an interesting geometric characteristic. The min and max aggregators return two farthest corners of the minimum bounding box specified by the set of user embeddings $\mathbf{U}_G$.

\subsection{Learning DeepGroup}
\label{sec:learning}
One can learn DeepGroup model parameters by the \emph{maximum likelihood estimation (MLE)} method. Given the observed groups $\calG$ and group-item interaction matrix $Y$, the log likelihood can be computed by
$$
\ell(\bftheta|\calG, \bfy) = \sum_{i=1}^l\sum_{j=1}^m y_{ij} \log \hat{y}_{ij} + (1-y_{ij}) \log \left(1-\hat{y}_{ij}\right),
$$
where $\hat{y}_{ij}=f(G_i,a_j|\bftheta)$ is DeepGroup's estimated probability for interaction of group $G_i$ with the item $a_j$. The maximum likelihood estimate of the model parameters are $$\hat{\bftheta}_{MLE} = \arg\max_{\bftheta} \ell(\bftheta|\calG, \bfy).$$ Equivalently, one can learn model parameters by minimizing loss $L(\bftheta|\calG, \bfy) = - \ell(\bftheta|\calG, \bfy)$. This loss function is the same as the \textit{binary cross-entropy loss}, which can be minimized by performing stochastic gradient descent (SGD) or any other optimization techniques.\footnote{While deploying the architecture of DeepGroup, this loss function can be modified to or extended with a \emph{pairwise loss} \cite{s_r+c_f+z_g+l_s+2012}, which directly optimizes the ranking of the group-item interactions. }

\section{Experiments}
\label{section:experiments}
We conduct extensive experiments to evaluate the effectiveness of our proposed DeepGroup model for both problems of group decision prediction and reverse social choice described in Section~\ref{section:problem}. We study the efficacy of DeepGroup by comparing its accuracy with some other baseline algorithms on four datasets.\footnote{The code for our experiments and DeepGroup can be found at \url{https://github.com/sarinasajadi/DeepGroup}}  

\subsection{Group Datasets}
The group datasets should consist of both group membership data (i.e., group structures) $\calG$, and group decisions in the form of group-item interaction matrices $\bfy$. Due to inaccessibility to such group datasets, we create our group datasets using real-world preference datasets, different group formation mechanisms, and group decision rules (or voting methods). 

\vskip 2mm
\noindent \textbf{Real-world preference datasets.}
As our problems (e.g., group decision and reverse social choice) are closely connected with social choice, we focus on preference rankings (or ordinal preferences) which are of special interest in social choice since they help circumvent the problem of interpersonal comparisons of ratings/utilities \cite{Arrow1963SocialChoice,sen70Collective}. We consider four real-world preference ranking datasets. Three datasets are from the 2002 Irish Election:\footnote{\url{http://www.preflib.org/data/election/irish/}} Dublin West with 9 candidates and 29,989 user preferences; Dublin North containing 43,942 user preferences over 12 candidates; and Meath containing 64,081 user preferences over 14 candidates. The user preferences in these datasets are partial rankings of the top-$t$ form (i.e., the ranking of the $t$ most preferred candidates). Our other dataset is the Sushi dataset consisting of $5000$ user preferences as complete rankings over $10$ varieties of sushi.\footnote{\url{http://www.preflib.org/data/election/sushi/}}

\vskip 2mm
\noindent \textbf{Group generation methods.}
To generate a set of groups $\calG$ from real-world preference datasets, we deploy various methods. The \textit{$\kappa$-participation group (KPG)} method first samples $n$ users from a preference dataset, then $\kappa$ times randomly partitions this set of users into size-constrained subsets (i.e., groups), whose sizes are bounded by $[s_{min},s_{max}]$. The KPG outputs the collection of all unique subsets generated by these $\kappa$ partitions. By varying $\kappa$, one can control the extent to which each user participated in different groups (or equivalently, the extent to which groups overlap with one another). 
 
The \textit{Random Similar Groups (RSG)} method randomly selects $l$ groups from a preference dataset, where the group size is randomly picked from $[s_{min},s_{max}]$ and group members have similar preferences. The preference similarity is enforced by ensuring that all pairwise Kendall-$\tau$ correlations of group members are at least $\tau_{sim}$. \textit{Random Dissimilar Groups (RDG)} method has the similar stochastic process of RSG with the difference that group members must have dissimilar preferences. The dissimilarity is imposed by ensuring that all pairwise Kendall-$\tau$ correlations of group members are at most $\tau_{dis}$.\footnote{RSG and RDG can be easily implemented by rejection sampling.} 
We set $s_{min}=2$, $s_{max}=10$, $\tau_{sim}=0.5$, and $\tau_{dis}=-0.5$ while varying other parameters.

\vskip 2mm
\noindent \textbf{Group decision rules.} 
We create the group-item interaction matrix $\bfy$ for each generated group set $\calG$ by voting rules \cite{compsoc16}. To do so, we aggregate user preferences of each group $G_i\in \calG$ to a group decision $a_j \in \calA$. We focus on Borda and plurality---two examples of positional scoring rules---in which an alternative $a$, for each preference ranking $r$, receives a score $g(a,r)$ based on its ranking position. Then, the group decision is the alternative with the highest cumulative score over all rankings of group members (ties are broken randomly). The Borda score function is $g_{B}(a, r) = m - r(a)$ and the plurality score function is $g_{P}(a, r) = \ind{r(a)=1}$ where $\ind{.}$ is an indicator function, and $r(a)$ represents the position of $a$ in the ranking $r$. In our experiments, for a fixed group set $\calG$, we either use Borda for all $G_i \in \calG$, use plurality for all $G_i \in \calG$, or uniformly at random select between Borda and plurality for each $G_i \in \calG$ (i.e., mixture of Borda and Plurality). All three preference aggregation strategies are unknown to the group recommendation methods studied in our experiments. The mixture strategy further challenges the group recommendation methods by stochastically diversifying the group decision rules among groups.    

\subsection{Benchmarks}
\label{sec:baselines}
We compare the prediction power of DeepGroup against some baseline methods. Due to the lack of any comparable algorithms for solving group decision prediction and social choice reverse problems, our baselines are either (i) adaptations of state-of-the-art deep learning methods to our problems (e.g. AGREE \cite{cao2018attentive}), or (ii) heuristics that we designed to provide reasonable benchmarks for DeepGroup.  
\vskip 1.5mm
\noindent \textit{AGREE} \cite{cao2018attentive}. This model adopts an attention mechanism in deep learning for group recommendations. It employs both user and group rating preferences for learning group preferences and consequently group recommendation. To make it comparable with DeepGroup for our problem, we deployed AGREE without user preferences while only inputting the groups and their top choices to this model. Except for this change, the other settings were left as default.\footnote{\url{https://github.com/LianHaiMiao/Attentive-Group-Recommendation}} 

\vskip 1.5mm
\noindent \textit{Popularity (Pop).} This method predicts the most popular group decisions in the training set as the group decision of any groups in the testing set. 

\vskip 1.5mm
\noindent \textit{Random Top Choice Plurality (RTCP).} For a group in the testing set, RTCP first guesses its users' top choices (or votes), then outputs the plurality winner (ties broken randomly). To guess user top choice, if a user belongs to at least one group in the training set, RTCP randomly picks one of its groups and assign that group's decision as its top choice. For those users that don't belong to any groups in the training set, the method guesses their top choices to be the popular group decision (or item) in the training set. 

\vskip 1.5mm
\noindent \textit{Overlap Similarity (O-Sim).} For a given group in the testing dataset, this method outputs the group decision of the most similar group in the training set, when the similarity is measured by the number of common members. 

\subsection{Experimental Setup}
\label{sec:expSetup}

In all our experiments, DeepGroup has four hidden layers (i.e., X=4) with 64, 32, 16, and 8 hidden units and the Relu activation function. To prevent overfitting, we use dropout over hidden layers with the probability of 0.8 for retaining each node in a hidden layer. The user and item embedding dimensions are both set to 64 and the Mean aggregator is used (unless noted otherwise). We optimize DeepGroup with Adam optimizer for 100 epochs with a learning rate of 0.001 and the batch size of 4096 (i.e., each mini-batch encompasses 4096 records of negative/positive group-user interactions along with group membership data).

For both group decision prediction and reverse social choice tasks, we compare the performance of DeepGroup and baselines by their prediction accuracy, measuring the percentage of correct top-choice prediction (i.e. group decision) for groups in the testing set. 

As our group dataset generation is stochastic, for each fixed setting (e.g., preference dataset, group set generation, group decision rule, etc.), we generate 20 instances of each group dataset setting and report our results as an average accuracy over those instances. In experiments focused on group decision prediction task, for each group dataset, we randomly selected 70\% of all groups and their group-item interactions as the training set and 30\% for the testing set. The groups in the testing sets are not observed in the training set, but their members might have appeared in some other groups in training.\footnote{DeepGroup sets the user embedding of a new user in the testing set with the average of learned user embeddings, thus assuming the average or \emph{default} preference.} When the task is the reverse social choice, we use each group dataset as the training set and create a testing set including the singleton groups of all users appeared in the groups of the training set. 

\begin{figure*}[tb]
\centering

\begin{tabular}{@{\hspace{-8pt}}c@{\hspace{-5pt}}c@{\hspace{-5pt}}c@{\hspace{-5pt}}c}
     \includegraphics[width=0.26\textwidth]{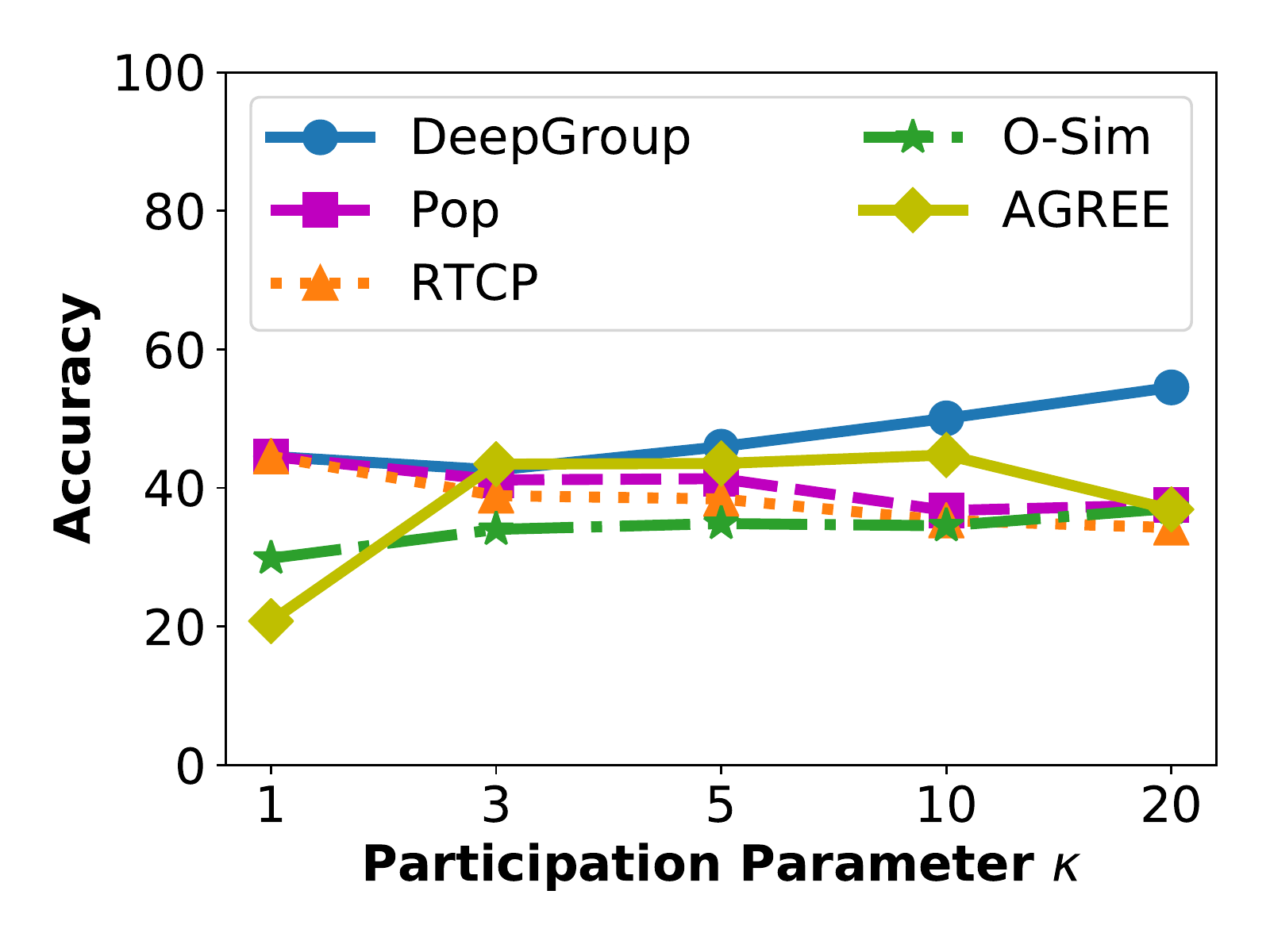}&
     \includegraphics[width=0.26\textwidth]{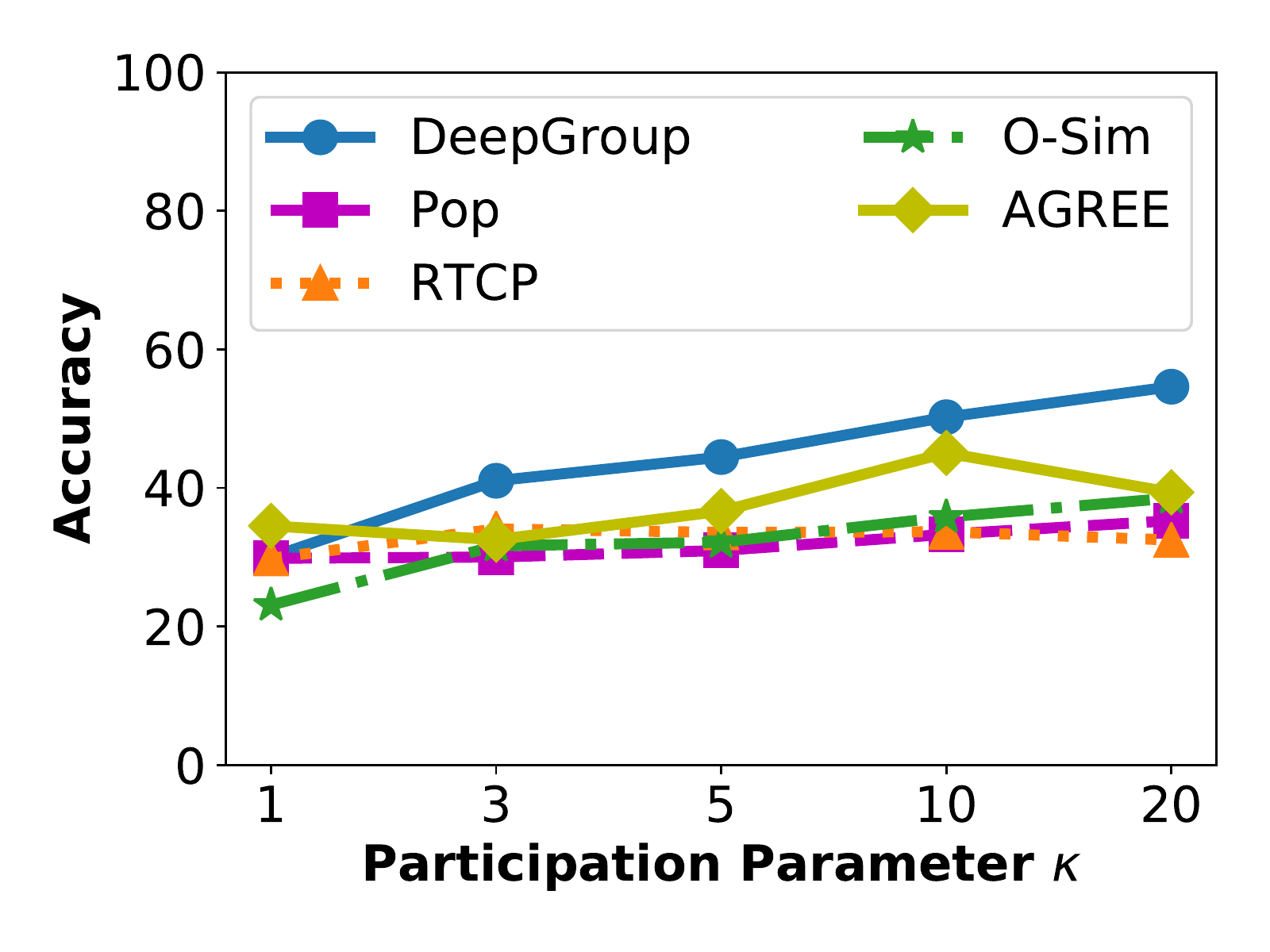} &
      \includegraphics[width=0.26\textwidth]{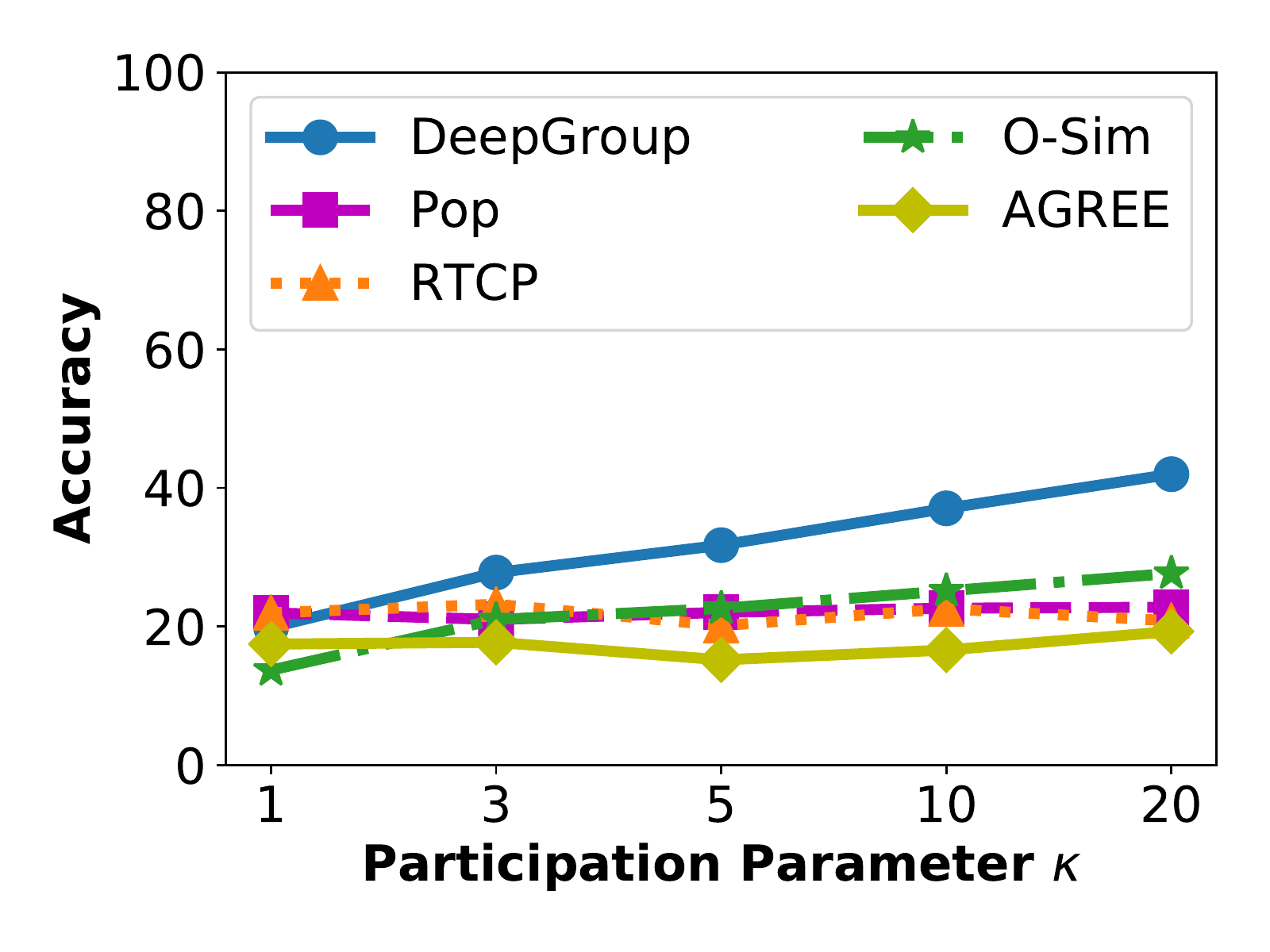}&
     \includegraphics[width=0.26\textwidth]{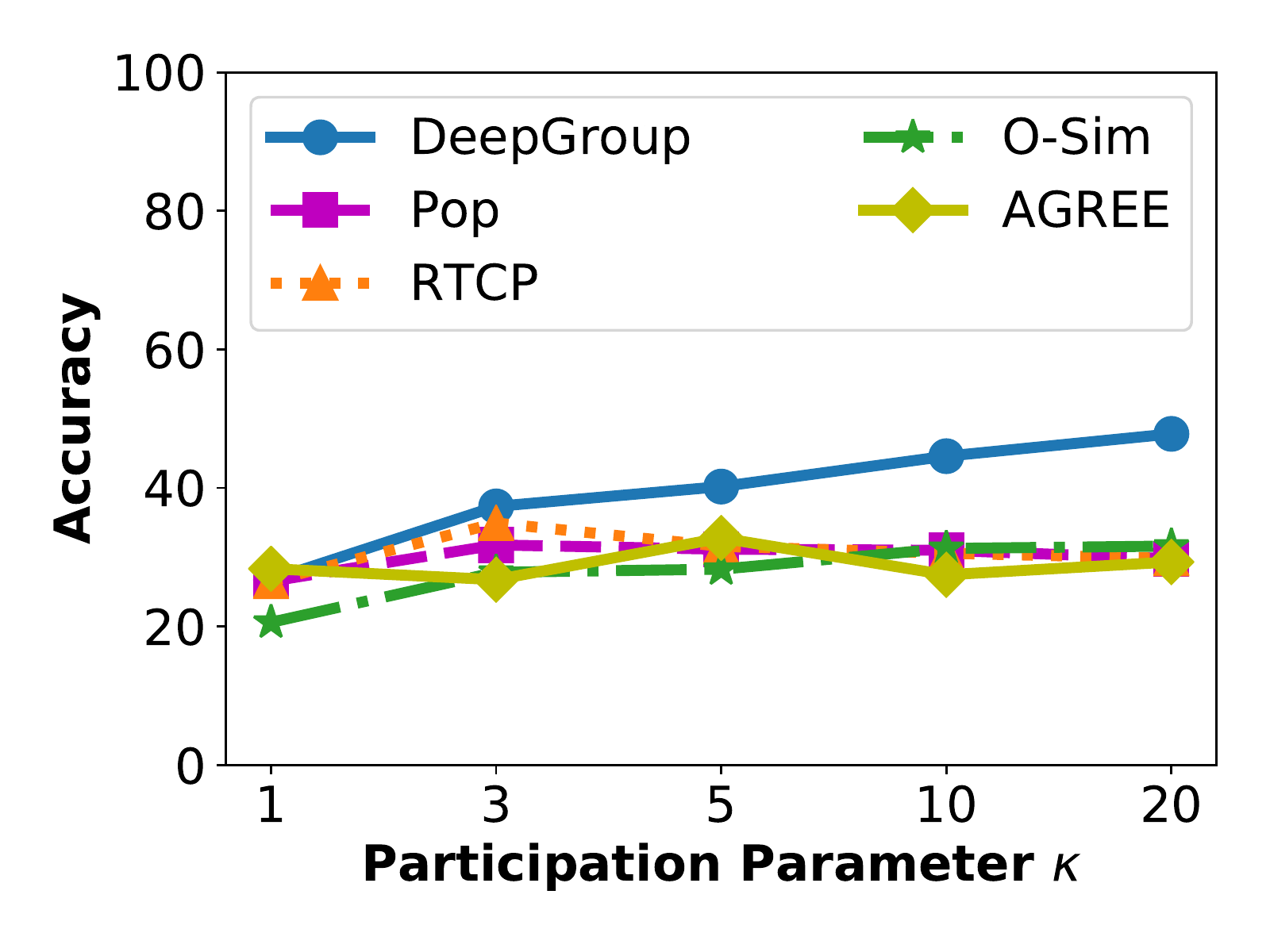}\\
       (a) Sushi &(b) Dublin West &(c) Dublin North& (d) Meath
\end{tabular}

 \caption{The accuracy of DeepGroup and other benchmarks for different group datasets generated on various preference datasets (a)--(d) with $\kappa$-participation method, and the plurality group decision rule.}
  \label{fig:overlap_plurality}
\end{figure*}

\begin{figure*}[tb]
\centering
\begin{tabular}{@{\hspace{-4pt}}c@{\hspace{-2pt}}c@{\hspace{-2pt}}c@{\hspace{-2pt}}c}
     \includegraphics[width=0.25\textwidth]{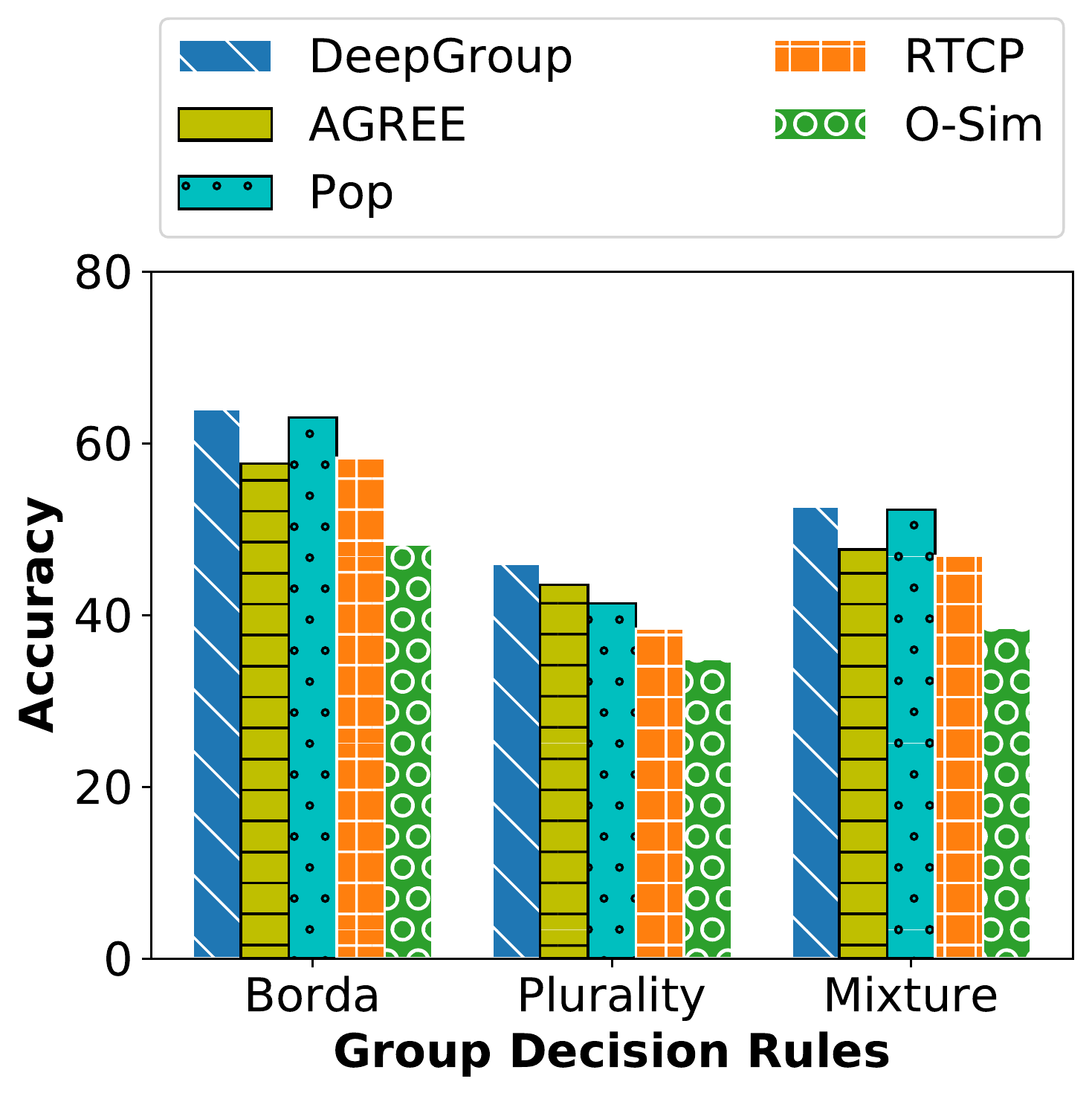}&
     \includegraphics[width=0.25\textwidth]{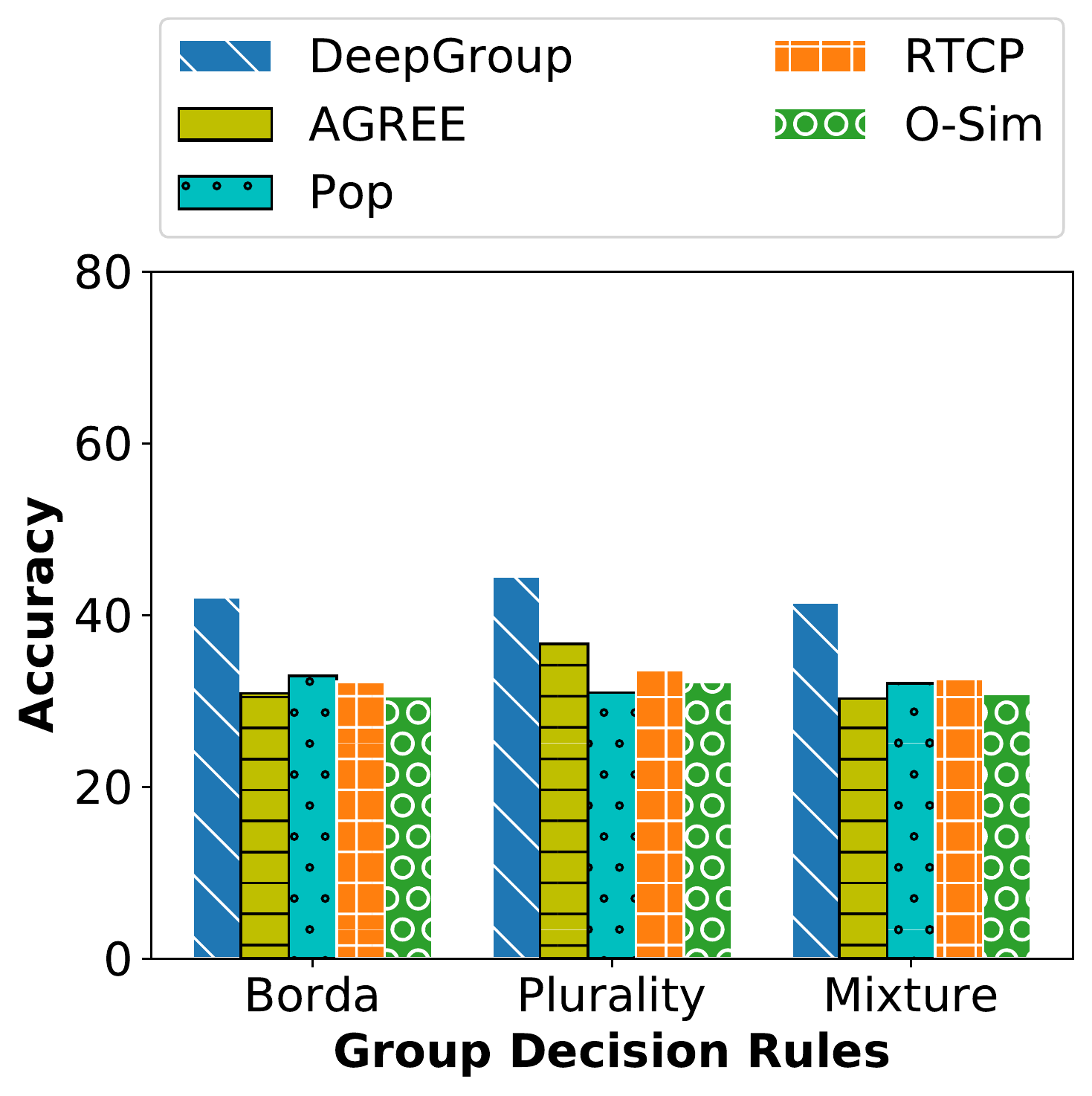}&
      \includegraphics[width=0.25\textwidth]{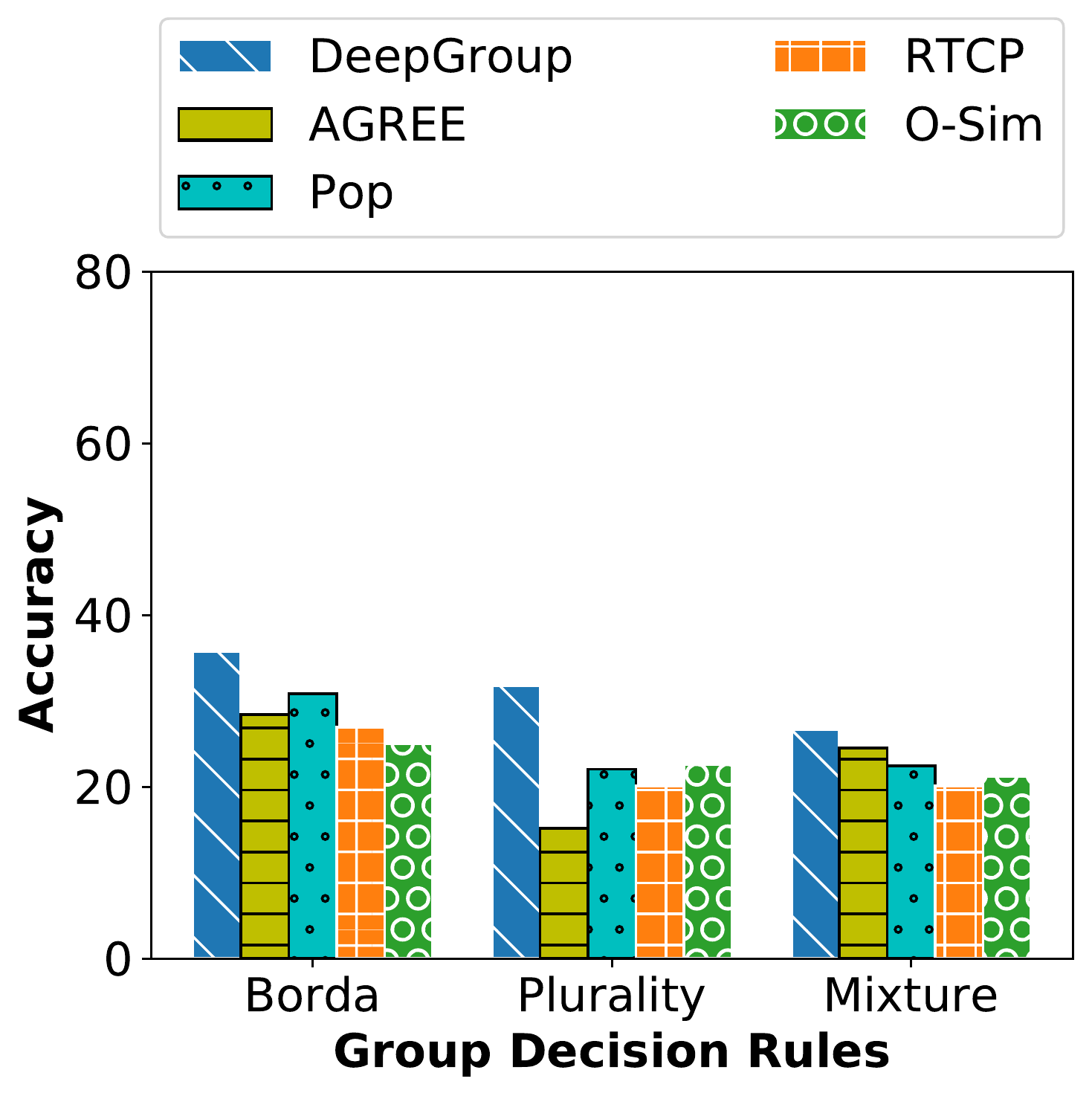}&
      \includegraphics[width=0.25\textwidth]{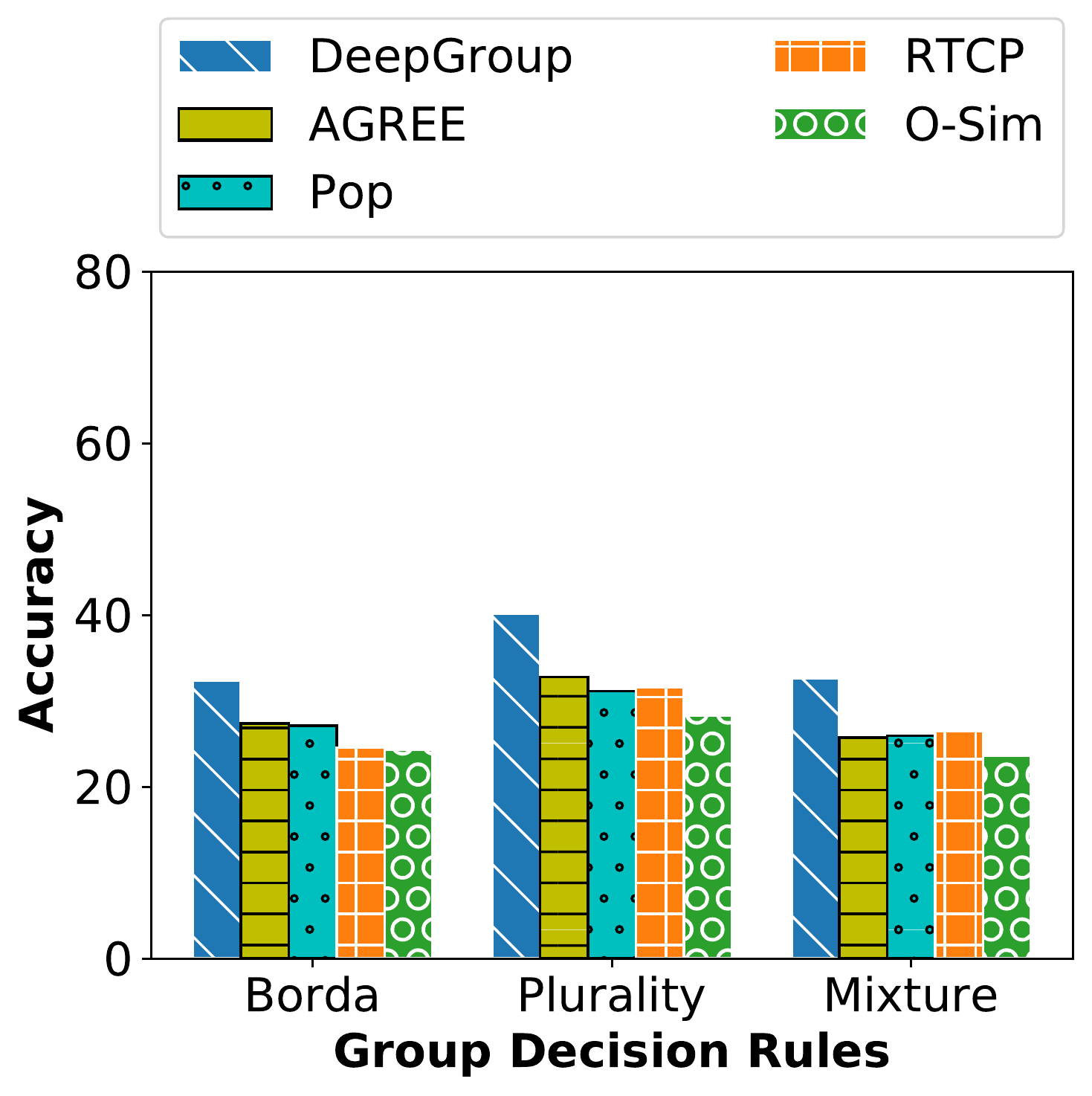}\\
      (a) Sushi &(b) Dublin West &(c) Dublin North& (d) Meath
\end{tabular}
 \caption{The accuracy of DeepGroup and other benchmarks over different group decision rules for different group datasets generated on different preference datasets (a)--(d) with $\kappa$-participation method (fixed $\kappa = 5$).}
  \label{fig:overlap_rules}
\end{figure*}

\subsection{Empirical Results}
We report the empirical results of our extensive experiments for both group decision prediction and reverse social choice. 

\vskip 2mm
\noindent \textbf{Group decision prediction on $\kappa$-participation group sets.}
We aim to investigate the effectiveness of DeepGroup in group decision prediction when compared to other benchmarks under various group datasets generated with real-world preference data and $\kappa$-participation group generation. We fix the group decision rule to plurality or Borda for all generated groups. By fixing the number of users $n=5000$ and varying $\kappa$ over $\{1,3,5,10,20\}$, we study how the performance of different methods change with more availability of implicit data (i.e., the participation of individuals in different group decisions). 

Figure \ref{fig:overlap_plurality} shows the accuracy of different methods for various group datasets for the plurality decision rule.\footnote{The results for Borda were qualitatively similar.} In all four datasets, DeepGroup performs comparably with benchmarks for $\kappa=1$ but outperforms the benchmarks for $\kappa \geq 3$. The performance of DeepGroup is more prominent as $\kappa$ increases (e.g., about 100\% improvement over the best baseline for $\kappa=20$ and Irish datasets). These results suggest that as users participate more in various group decision-making processes, the model can more accurately learn their embeddings and consequently the embeddings of their groups. We observe that none of the benchmarks except AGREE exhibit the same behavior since their performances remain almost steady as $\kappa$ increases. Although the accuracy of AGREE improves as $\kappa$ increases, its accuracy is still not comparable to DeepGroup.

\begin{figure*}[tb]
\centering
\begin{tabular}{@{\hspace{-8pt}}c@{\hspace{-4pt}}c@{\hspace{-4pt}}c@{\hspace{-4pt}}c}
     \includegraphics[width=0.26\textwidth]{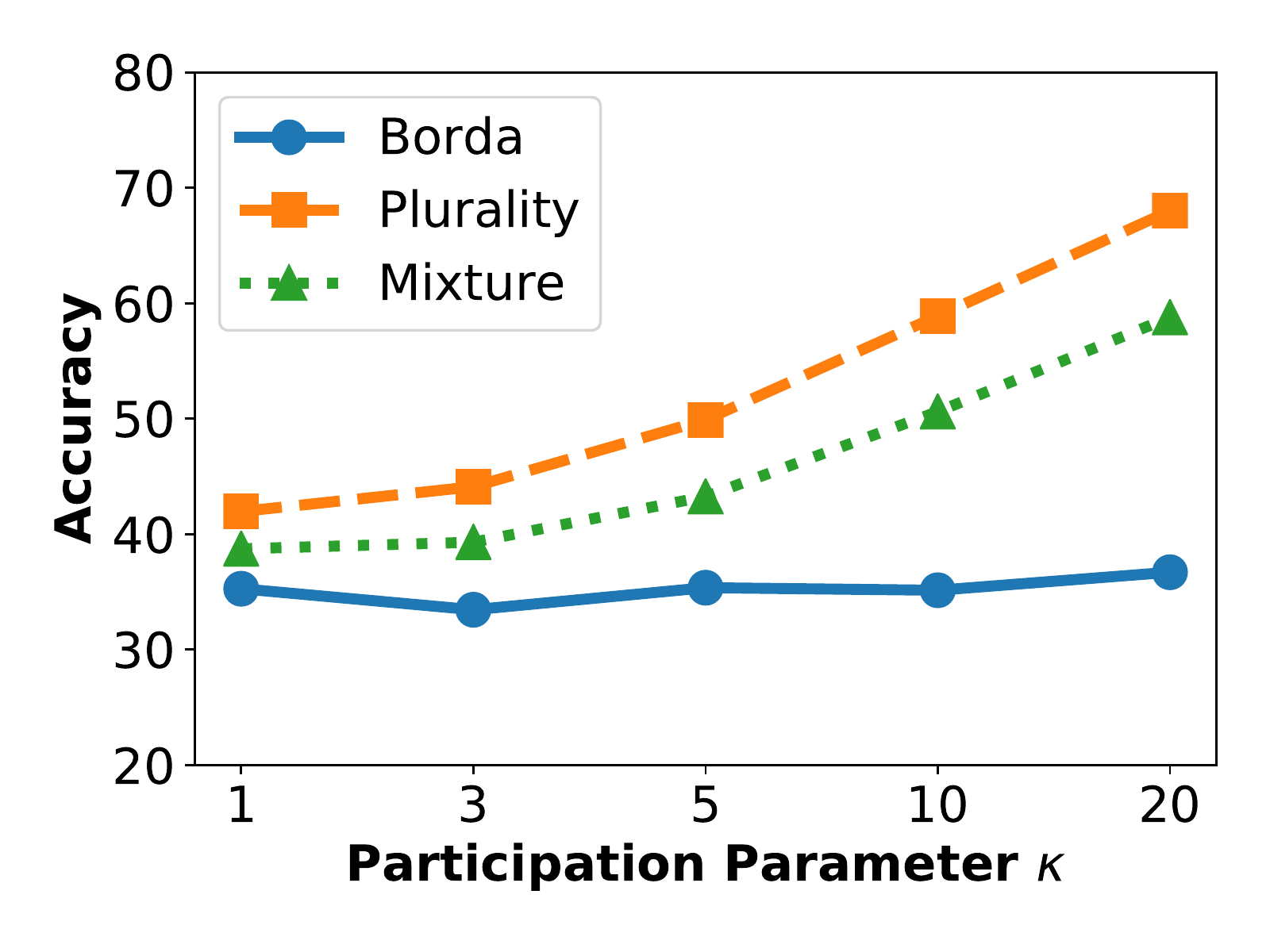}&
     \includegraphics[width=0.26\textwidth]{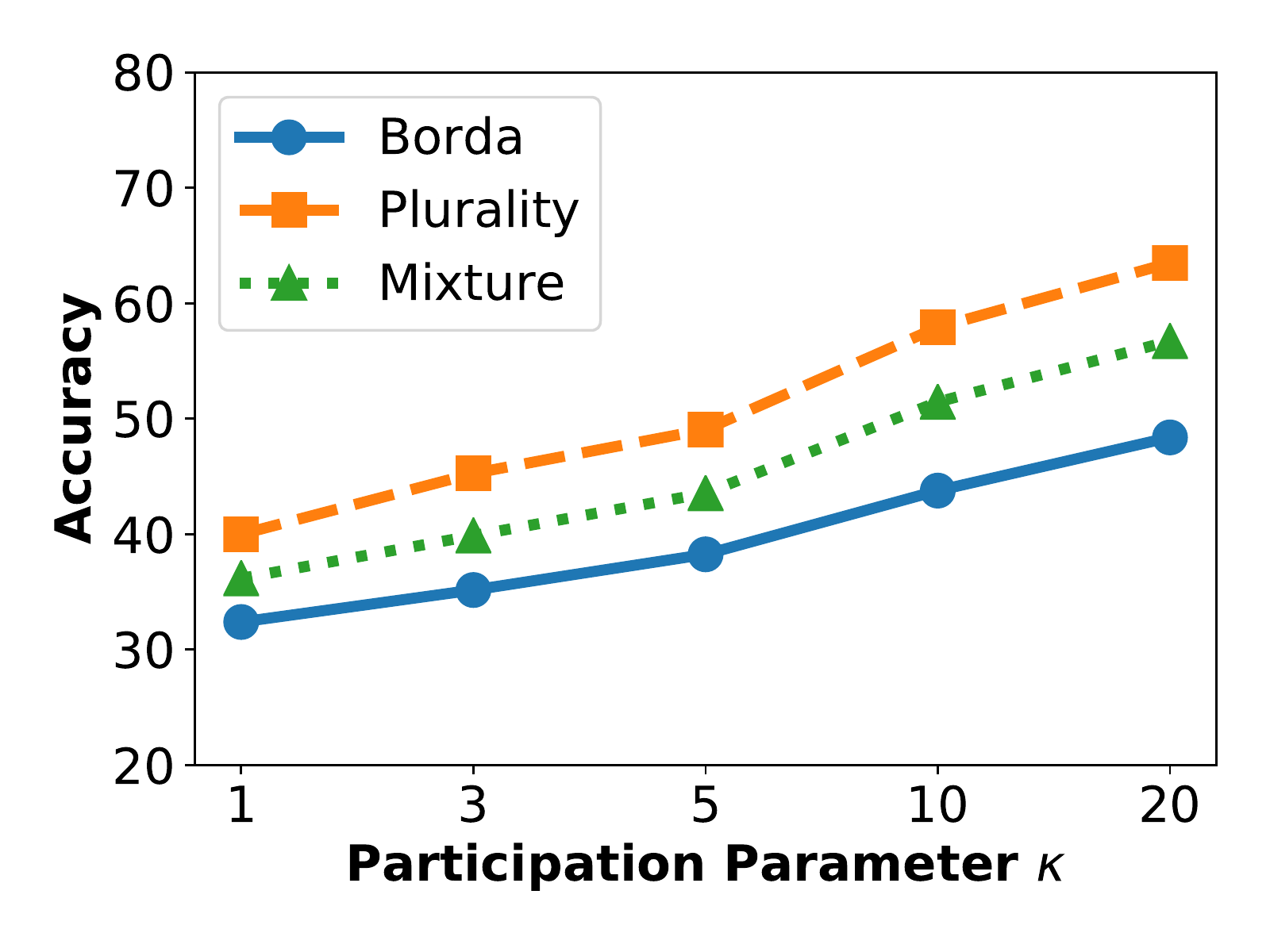}&
      \includegraphics[width=0.26\textwidth]{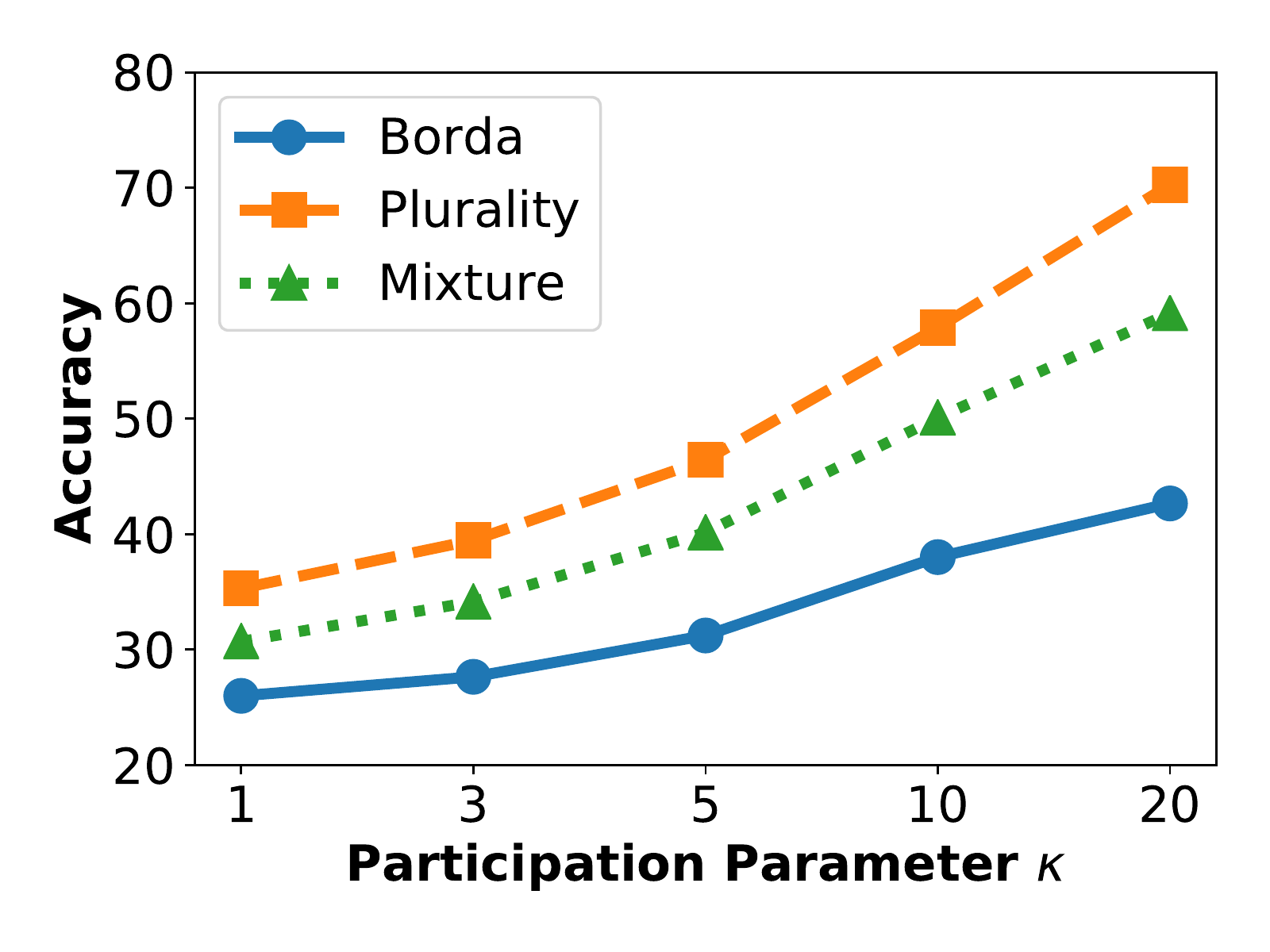}&
      \includegraphics[width=0.26\textwidth]{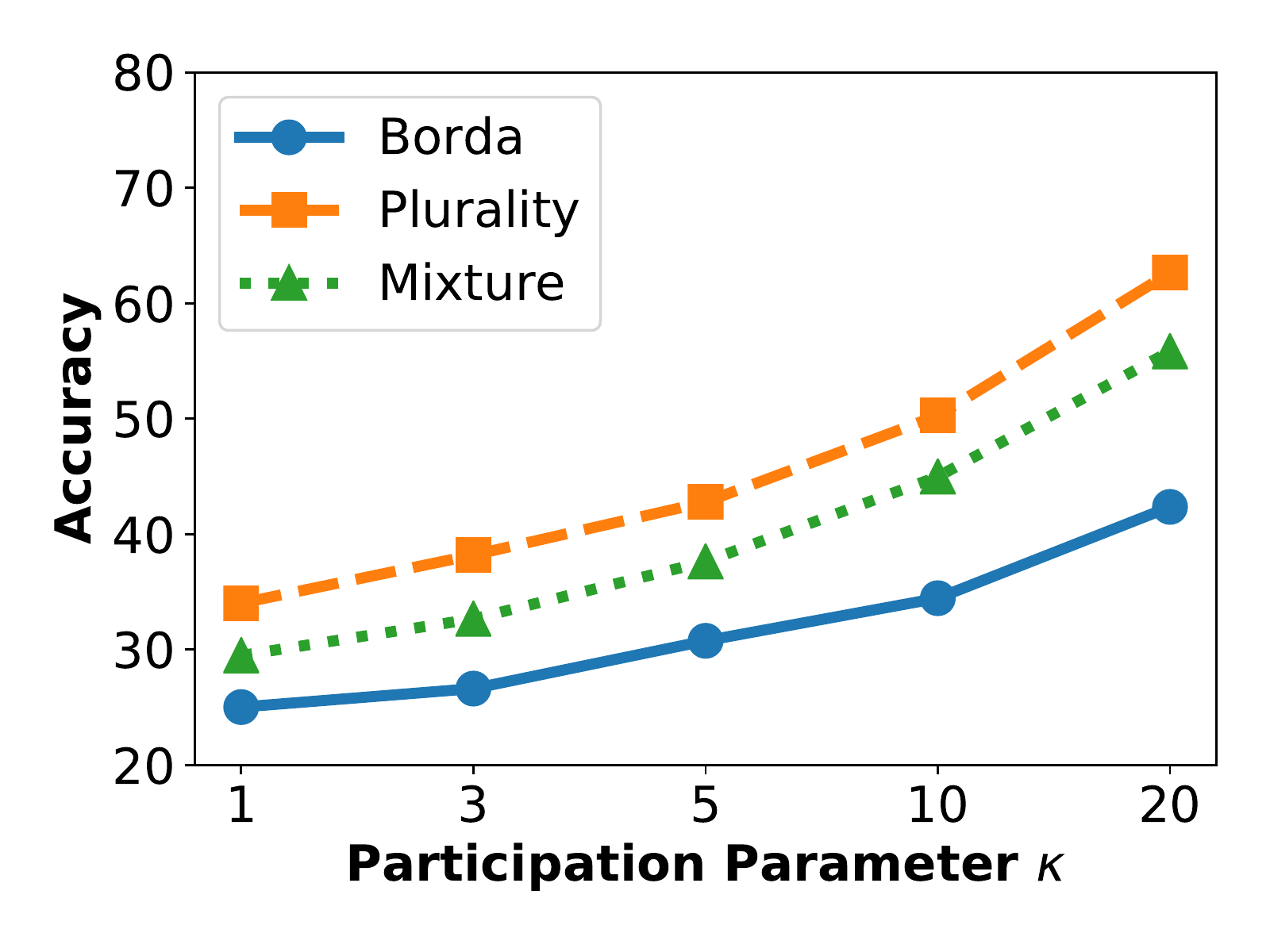}\\
      (a) Sushi &(b) Dublin West & (c) Dublin North& (d) Meath
\end{tabular}
 \caption{The accuracy of DeepGroup for reverse social choice, group datasets generated by different group decision rules on various preference datasets (a)--(d) with $\kappa$-participation method.}
  \label{fig:reverse_rules_compare}
\end{figure*}

\begin{figure*}[tb]
\centering
\begin{tabular}{@{\hspace{-1pt}}c@{\hspace{-1pt}}c@{\hspace{-1pt}}c@{\hspace{-1pt}}c}
     \includegraphics[width=0.25\textwidth]{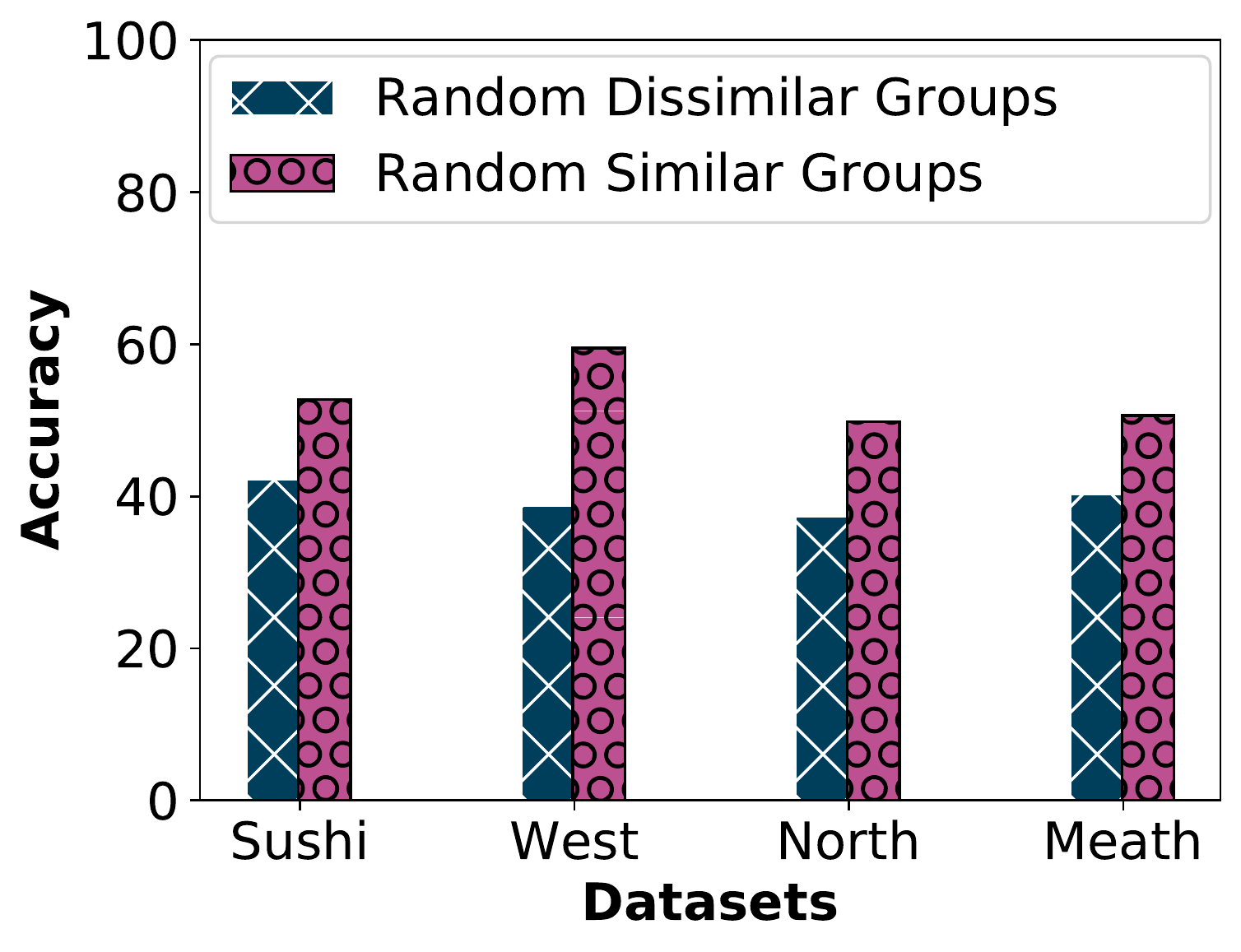}&
     \includegraphics[width=0.25\textwidth]{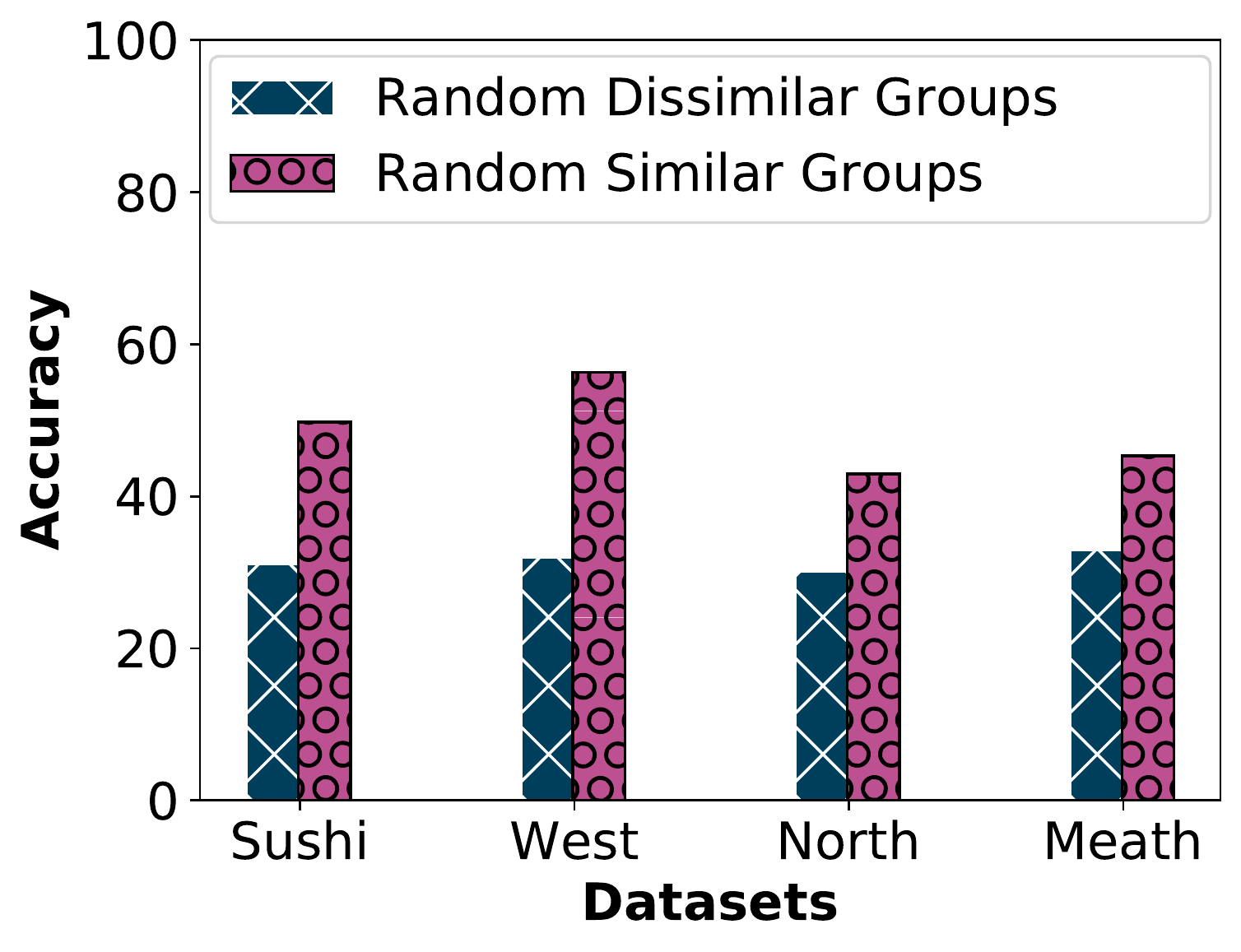}&
      \includegraphics[width=0.25\textwidth]{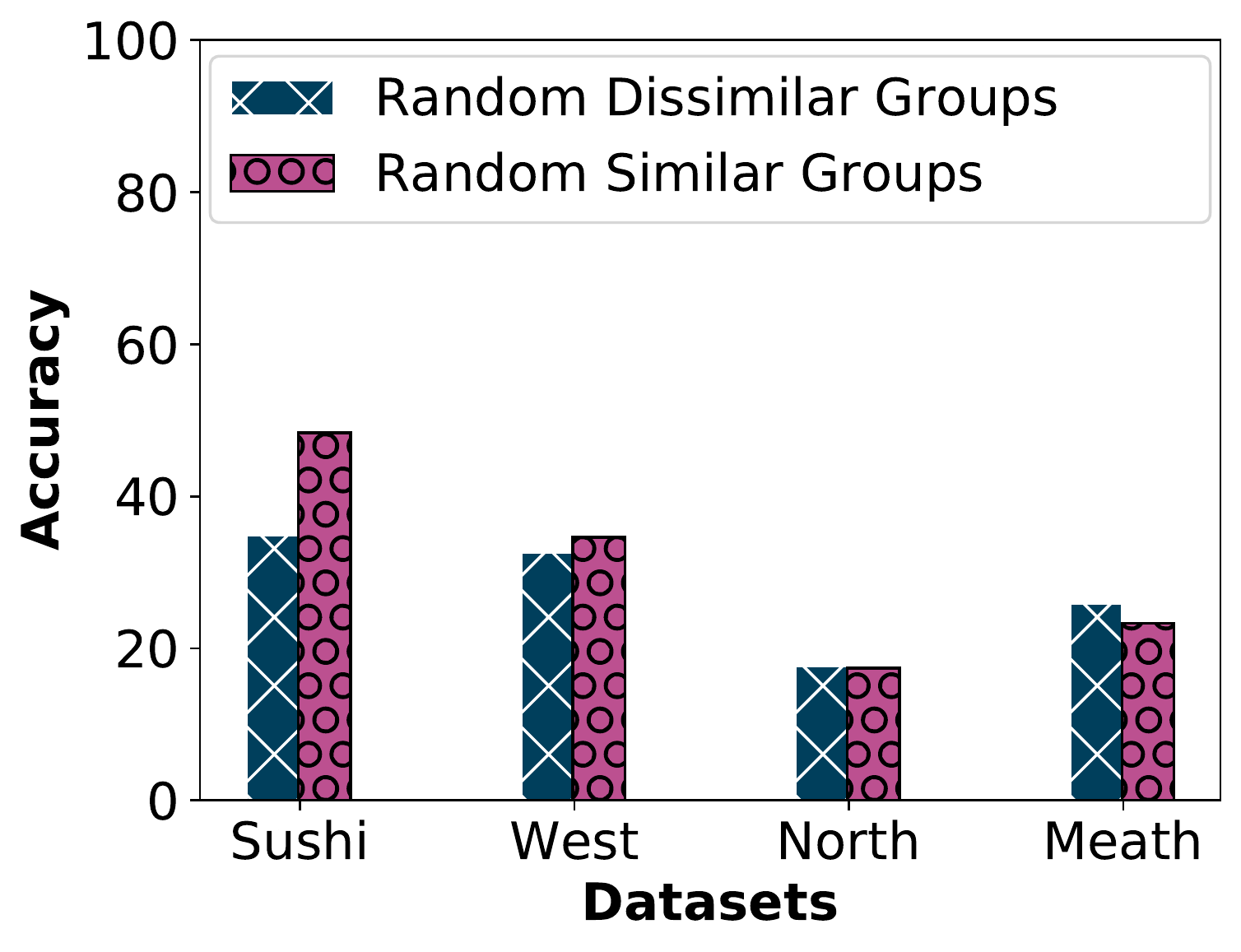}&
      \includegraphics[width=0.25\textwidth]{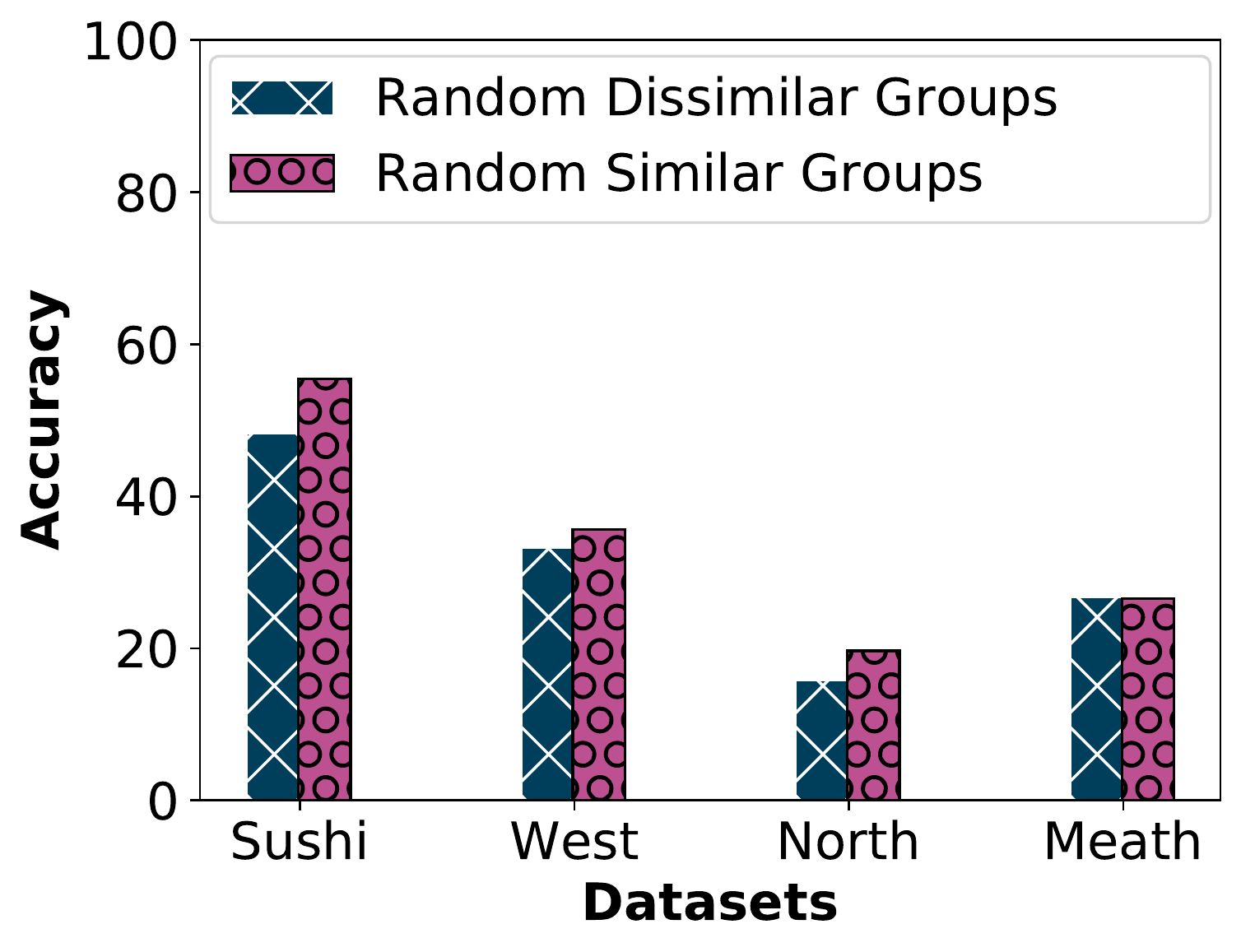}\\
     (a) Rev. Social Choice, Plurality &(b) Rev. Social Choice, Borda & (c) Group Decision, Plurality& (d) Group Decision, Borda
\end{tabular}
 \caption{The accuracy of DeepGroup for similar and dissimilar random groups,  group decision prediction and reverse social choice tasks, and Borda and Plurality group decision rules (a)--(d).}
  \label{fig:RG_compare_all}
\end{figure*}
\vskip 1mm
\noindent \textbf{The effect of decision rule on group decision prediction.}
To investigate the effect of different group decision rules on group decision prediction, we compare the accuracy of DeepGroup and baselines on group datasets generated with various decision rules. We use $\kappa$-participation group generation method with fixed $\kappa=5$ and $n=5000$. Figure~\ref{fig:overlap_rules} shows the accuracy of methods for various group decision rules (i.e.,  Borda, Plurality, and their mixtures) over different preference datasets.
For all preference datasets, DeepGroup outperforms others over all decision rules to the various extent. It seems that DeepGroup offers the most improvement over baselines for plurality and the least improvement for Borda. One interesting observation is that DeepGroup still performs fairly well for the mixture of Borda and Plurality. This suggests that (a) DeepGroup does not necessarily require to be aware of group decision rules for successful prediction, and (b) DeepGroup can perform well when different groups use inconsistent decision rules.   

\vskip 1mm
\noindent \textbf{Reverse social choice and group decision rules.}
We study the accuracy of DeepGroup for reverse social choice (i.e., predicting individual preferences of group members) when various group decision rules are applied. We use $\kappa$-participation while varying $\kappa$ over $\{1,3,5,10,20\}$. Figure \ref{fig:reverse_rules_compare} shows the accuracy of DeepGroup for various group decisions rules, preference datasets, and $\kappa$ parameter. For all group decision rules and preference datasets, the accuracy of DeepGroup increases with participation factor $\kappa$. This implies that user personal preferences can be predicted more accurately if users participate in more group decisions.\footnote{We observed similar patterns for group decision prediction task.}  One can observe an interesting pattern by comparing DeepGroup accuracy over group decision rules.  For the Plurality decision rule, the accuracy is always the highest in all the datasets whereas Borda has the lowest accuracy. This observation is surprising: despite requiring the least preference data (i.e., only top choice) for decision making, plurality has the highest privacy leakage as the personal preferences can be predicted more accurately when it is deployed. In contrast, Borda has the lowest privacy leakage in this sense.  Another important observation emerges from this experiment: when the decision rule is not inconsistent among the groups in a dataset (e.g., the mixture of plurality and Borda), DeepGroup is still effective in predicting the individual preferences.

\vskip 2mm
\noindent \textbf{The effect of homophily and heterophily on DeepGroup.}
We aim to explore how the performance of DeepGroup changes when the group members possess similar or dissimilar preferences (homophily vs heterophily). To this end, we study the accuracy of DeepGroup for both group decision prediction and social reverse choice tasks, when the groups are generated by either of Random Similar Group (RSG) or Random Dissimilar Group (RDG). We fix the number of randomly generated groups $l=1000$ for this experiment. Figure~\ref{fig:RG_compare_all} shows the accuracy of DeepGroup for various group generation methods, both prediction tasks, Borda and Plurality group decision rules, and different preference datasets. For both reverse social choice and group decision prediction, DeepGroup has higher performance for homophilic groups (i.e. group members have similar preferences) compared to heterophilic groups (i.e., groups with dissimilar preferences) regardless of the underlying group decision rule. From the privacy perspective, this result (especially for reverse social choice) implies that privacy-leakage of user preferences is the highest when the groups are composed of like-minded individuals. The rationale behind this observation is that a group's revealed decision is a good representative of all its group members' preferences when they are like-minded.

\section{Conclusion and Future Work}
\label{section:conclusion}
We formulate the problem of group recommendation from group implicit feedback, with the goal of making item recommendation to a new group of users in the absence of personal user preferences. To address this problem, we introduce DeepGroup--- a novel model for learning group representation based on observed group-item interactions. We conduct an extensive series of experiments to evaluate DeepGroup accuracy over various datasets and benchmarks while focusing on two special instances of our problem, reverse social choice and group decision prediction. Our findings confirm the effectiveness of DeepGroup in addressing these two problems. Our empirical results also show that different group decision rules (e.g., plurality, Borda, etc.) exhibit privacy leakage of concealed personal preferences with the various extent. Surprisingly, plurality, despite requiring less information than Borda, suffers more privacy leakage than Borda.

There are many fascinating directions to explore in future work. One can theoretically analyze some well-known voting rules in the context of our reverse social choice problem. These analyses can shed light on privacy-preserving characteristics of voting rules when the group decisions are publicly announced.  Our DeepGroup model can possibly be improved by incorporating ranking loss functions and deploying more complex latent aggregator functions.  DeepGroup is also a building block for the broader investigation of deep learning methods for group recommendation with group implicit feedback. Of practical importance is to extend the model with group and item features (e.g., descriptions, demographic information, etc.), side information (e.g., social networks between users), or context (e.g., time, location, etc.).
%
%
%
\bibliographystyle{IEEEtran}
\bibliography{sample,RecSys,DecisionSocial}

\begin{thebibliography}{10}
\providecommand{\url}[1]{#1}
\csname url@samestyle\endcsname
\providecommand{\newblock}{\relax}
\providecommand{\bibinfo}[2]{#2}
\providecommand{\BIBentrySTDinterwordspacing}{\spaceskip=0pt\relax}
\providecommand{\BIBentryALTinterwordstretchfactor}{4}
\providecommand{\BIBentryALTinterwordspacing}{\spaceskip=\fontdimen2\font plus
\BIBentryALTinterwordstretchfactor\fontdimen3\font minus
  \fontdimen4\font\relax}
\providecommand{\BIBforeignlanguage}[2]{{%
\expandafter\ifx\csname l@#1\endcsname\relax
\typeout{** WARNING: IEEEtran.bst: No hyphenation pattern has been}%
\typeout{** loaded for the language `#1'. Using the pattern for}%
\typeout{** the default language instead.}%
\else
\language=\csname l@#1\endcsname
\fi
#2}}
\providecommand{\BIBdecl}{\relax}
\BIBdecl

\bibitem{mccarthy2006cats}
K.~McCarthy, M.~Salam{\'o}, L.~Coyle, L.~McGinty, B.~Smyth, and P.~Nixon,
  ``Cats: A synchronous approach to collaborative group recommendation,'' in
  \emph{Florida Artificial Intelligence Research Society Conference (FLAIRS)},
  ser. FLAIRS'06, 2006, pp. 86--91.

\bibitem{crossen2002flytrap}
A.~Crossen, J.~Budzik, and K.~J. Hammond, ``Flytrap: Intelligent group music
  recommendation,'' in \emph{Proceedings of the 7th international conference on
  Intelligent user interfaces}, ser. IUI'02, 2002, pp. 184--185.

\bibitem{rakesh2016probabilistic}
V.~Rakesh, W.-C. Lee, and C.~K. Reddy, ``Probabilistic group recommendation
  model for crowdfunding domains,'' in \emph{Proceedings of the Ninth ACM
  International Conference on Web Search and Data Mining}, ser. WSDM'16, 2016,
  pp. 257--266.

\bibitem{pizzutilo2005group}
S.~Pizzutilo, B.~De~Carolis, G.~Cozzolongo, and F.~Ambruoso, ``Group modeling
  in a public space: methods, techniques, experiences,'' in \emph{Proceedings
  of the 5th WSEAS International Conference on Applied Informatics and
  Communications}, ser. AIC'05, 2005, pp. 175--180.

\bibitem{yu2006tv}
Z.~Yu, X.~Zhou, Y.~Hao, and J.~Gu, ``Tv program recommendation for multiple
  viewers based on user profile merging,'' \emph{User Modeling and User-Adapted
  Interaction}, vol.~16, no.~1, pp. 63--82, 2006.

\bibitem{o2001polylens}
M.~O'connor, D.~Cosley, J.~A. Konstan, and J.~Riedl, ``Polylens: a recommender
  system for groups of users,'' ser. ECSCW'01, 2001, pp. 199--218.

\bibitem{BMR2010group}
L.~Baltrunas, T.~Makcinskas, and F.~Ricci, ``Group recommendations with rank
  aggregation and collaborative filtering,'' in \emph{Proceedings of the 4th
  ACM Conference on Recommender Systems}, ser. RecSys'10, 2010, pp. 119--126.

\bibitem{Amer-Yahia:2009}
S.~Amer-Yahia, S.~B. Roy, A.~Chawlat, G.~Das, and C.~Yu, ``Group
  recommendation: Semantics and efficiency,'' \emph{Proc. VLDB Endow.}, vol.~2,
  no.~1, pp. 754--765, 2009.

\bibitem{SYMM2011}
S.~Seko, T.~Yagi, M.~Motegi, and S.~Muto, ``Group recommendation using feature
  space representing behavioral tendency and power balance among members,'' in
  \emph{Proceedings of the Fifth ACM Conference on Recommender Systems}, ser.
  RecSys'11, 2011, pp. 101--108.

\bibitem{GXLBH2010}
M.~Gartrell, X.~Xing, Q.~Lv, A.~Beach, R.~Han, S.~Mishra, and K.~Seada,
  ``Enhancing group recommendation by incorporating social relationship
  interactions,'' in \emph{Proceedings of the 16th ACM International Conference
  on Supporting Group Work}, ser. GROUP'10, 2010, pp. 97--106.

\bibitem{POSN2015}
A.~Salehi-Abari and C.~Boutilier, ``Preference-oriented social networks: Group
  recommendation and inference,'' in \emph{Proceedings of the 9th ACM
  Conference on Recommender Systems (RecSys-15)}, ser. RecSys'15, 2015.

\bibitem{XiaoRecsys2017}
L.~Xiao, Z.~Min, Z.~Yongfeng, G.~Zhaoquan, L.~Yiqun, and M.~Shaoping,
  ``Fairness-aware group recommendation with pareto-efficiency,'' in
  \emph{Proceedings of the Eleventh ACM Conference on Recommender Systems},
  ser. RecSys'17, 2017, pp. 107--115.

\bibitem{cao2018attentive}
D.~Cao, X.~He, L.~Miao, Y.~An, C.~Yang, and R.~Hong, ``Attentive group
  recommendation,'' in \emph{The 41st International ACM SIGIR Conference on
  Research \& Development in Information Retrieval}, 2018, pp. 645--654.

\bibitem{kalech-elicitation:2011}
M.~Kalech, S.~Kraus, G.~A. Kaminka, and C.~V. Goldman, ``Practical voting rules
  with partial information,'' \emph{Journal of Autonomous Agents and
  Multi-Agent Systems}, vol.~22, no.~1, pp. 151--182, 2011.

\bibitem{Xia-Conitzer:aaai08}
L.~Xia and V.~Conitzer, ``Determining possible and necessary winners under
  common voting rules given partial orders,'' Chicago, 2008, pp. 202--207.

\bibitem{Lu-Boutilier_elicitation:ijcai11}
T.~Lu and C.~Boutilier, ``Robust approximation and incremental elicitation in
  voting protocols,'' in \emph{Proceedings of the Twenty-Second International
  Joint Conference on Artificial Intelligence}, ser. IJCAI'11, Barcelona, 2011,
  pp. 287--293.

\bibitem{hughes2015computing}
D.~Hughes, K.~Hwang, and L.~Xia, ``Computing optimal bayesian decisions for
  rank aggregation via mcmc sampling,'' in \emph{UAI}, 2015, pp. 385--394.

\bibitem{doucette2015conventional}
J.~A. Doucette, K.~Larson, and R.~Cohen, ``Conventional machine learning for
  social choice,'' in \emph{Proceedings of the Twenty-Ninth AAAI Conference on
  Artificial Intelligence}, ser. AAAI'15.\hskip 1em plus 0.5em minus
  0.4em\relax AAAI Press, 2015, pp. 858--864.

\bibitem{Masthoff2004}
J.~Masthoff, ``Group modeling: Selecting a sequence of television items to suit
  a group of viewers,'' \emph{User Modeling and User-Adapted Interaction},
  vol.~14, no.~1, pp. 37--85, 2004.

\bibitem{compsoc16}
F.~Brandt, V.~Conitzer, U.~Endriss, J.~Lang, and A.~Procaccia, Eds.,
  \emph{Handbook of Computational Social Choice}.\hskip 1em plus 0.5em minus
  0.4em\relax Cambridge University Press, 2016.

\bibitem{felfernig2018group}
A.~Felfernig, L.~Boratto, M.~Stettinger, and M.~Tkal{\v{c}}i{\v{c}},
  \emph{Group Recommender Systems: An introduction}.\hskip 1em plus 0.5em minus
  0.4em\relax Springer, 2018.

\bibitem{yin2019social}
H.~Yin, Q.~Wang, K.~Zheng, Z.~Li, J.~Yang, and X.~Zhou, ``Social
  influence-based group representation learning for group recommendation,'' in
  \emph{2019 IEEE 35th International Conference on Data Engineering
  (ICDE)}.\hskip 1em plus 0.5em minus 0.4em\relax IEEE, 2019, pp. 566--577.

\bibitem{huang2020efficient}
Z.~Huang, X.~Xu, H.~Zhu, and M.~Zhou, ``An efficient group recommendation model
  with multiattention-based neural networks,'' \emph{IEEE Transactions on
  Neural Networks and Learning Systems}, vol.~31, 2020.

\bibitem{hu2014}
L.~Hu, J.~Cao, G.~Xu, L.~Cao, Z.~Gu, and W.~Cao, ``Deep modeling of group
  preferences for group-based recommendation,'' in \emph{Proceedings of the
  Twenty-Eighth AAAI Conference on Artificial Intelligence}, ser.
  AAAI'14.\hskip 1em plus 0.5em minus 0.4em\relax AAAI Press, 2014, pp.
  1861--1867.

\bibitem{dara2020survey}
S.~Dara, C.~R. Chowdary, and C.~Kumar, ``A survey on group recommender
  systems,'' \emph{Journal of Intelligent Information Systems}, vol.~54, pp.
  271--295, 2020.

\bibitem{mccarthy1998musicfx}
J.~F. McCarthy and T.~D. Anagnost, ``Musicfx: an arbiter of group preferences
  for computer supported collaborative workouts,'' in \emph{Proceedings of the
  1998 ACM Conference on Computer Supported Cooperative Work}, ser. CSCW'98,
  1998, pp. 363--372.

\bibitem{kim2010group}
J.~K. Kim, H.~K. Kim, H.~Y. Oh, and Y.~U. Ryu, ``A group recommendation system
  for online communities,'' \emph{International Journal of Information
  Management}, vol.~30, no.~3, pp. 212--219, 2010.

\bibitem{BF2010Rec}
S.~Berkovsky and J.~Freyne, ``Group-based recipe recommendations: Analysis of
  data aggregation strategies,'' in \emph{Proceedings of the 4th ACM Conference
  on Recommender Systems}, ser. RecSys'10, 2010, pp. 111--118.

\bibitem{GroupIM2020}
A.~Sankar, Y.~Wu, Y.~Wu, W.~Zhang, H.~Yang, and H.~Sundaram, ``Groupim: A
  mutual information maximization framework for neural group recommendation,''
  in \emph{Proceedings of the 43rd International ACM SIGIR Conference on
  Research and Development in Information Retrieval}, ser. SIGIR '20, 2020, p.
  1279–1288.

\bibitem{konczak-lang:preferences05}
K.~Konczak and J.~Lang, ``Voting procedures with incomplete preferences,'' in
  \emph{IJCAI-05 Workshop on Advances in Preference Handling}, Edinburgh, 2005,
  pp. 124--129.

\bibitem{Lu-Boutilier_multiwinner:ijcai13}
T.~Lu and C.~Boutilier, ``Multi-winner social choice with incomplete
  preferences,'' in \emph{Proceedings of the Twenty-Third International Joint
  Conference on Artificial Intelligence}, ser. IJCAI'13, Beijing, 2013.

\bibitem{ZYST2019}
S.~Zhang, L.~Yao, A.~Sun, and Y.~Tay, ``Deep learning based recommender system:
  A survey and new perspectives,'' \emph{ACM Computing Surveys (CSUR)},
  vol.~52, no.~1, pp. 1--38, 2019.

\bibitem{wideDeep2016}
H.-T. Cheng, L.~Koc, J.~Harmsen, T.~Shaked, T.~Chandra, H.~Aradhye,
  G.~Anderson, G.~Corrado, W.~Chai, M.~Ispir, R.~Anil, Z.~Haque, L.~Hong,
  V.~Jain, X.~Liu, and H.~Shah, ``Wide \& deep learning for recommender
  systems,'' in \emph{Proceedings of the 1st Workshop on Deep Learning for
  Recommender Systems}, 2016, pp. 7--10.

\bibitem{he2017neural}
X.~He, L.~Liao, H.~Zhang, L.~Nie, X.~Hu, and T.-S. Chua, ``Neural collaborative
  filtering,'' in \emph{Proceedings of the 26th International Conference on
  World Wide Web}, 2017, pp. 173--182.

\bibitem{wu2017recurrent}
C.-Y. Wu, A.~Ahmed, A.~Beutel, A.~J. Smola, and H.~Jing, ``Recurrent
  recommender networks,'' in \emph{Proceedings of the 10th ACM International
  Conference on Web Search and Data Mining}, 2017, pp. 495--503.

\bibitem{ZWC-JCA-2019}
Z.~Zhu, J.~Wang, and J.~Caverlee, ``Improving top-k recommendation via joint
  collaborative autoencoders,'' in \emph{The World Wide Web Conference}, 2019,
  pp. 3483--3482.

\bibitem{wu2016collaborative}
Y.~Wu, C.~DuBois, A.~X. Zheng, and M.~Ester, ``Collaborative denoising
  auto-encoders for top-n recommender systems,'' in \emph{Proceedings of the
  Ninth ACM International Conference on Web Search and Data Mining}, 2016, pp.
  153--162.

\bibitem{LS-2017-KDD}
X.~Li and J.~She, ``Collaborative variational autoencoder for recommender
  systems,'' in \emph{Proceedings of the 23rd ACM SIGKDD International
  Conference on Knowledge Discovery and Data Mining}, 2017, pp. 305--314.

\bibitem{liang2018variational}
D.~Liang, R.~G. Krishnan, M.~D. Hoffman, and T.~Jebara, ``Variational
  autoencoders for collaborative filtering,'' in \emph{Proceedings of the 27th
  International World Wide Web Conference}, 2018, pp. 689--698.

\bibitem{askari2020joint}
B.~Askari, J.~Szlichta, and A.~Salehi-Abari, ``Joint variational autoencoders
  for recommendation with implicit feedback,'' \emph{arXiv preprint
  arXiv:2008.07577}, 2020.

\bibitem{delic2017comprehensive}
A.~Delic and J.~Neidhardt, ``A comprehensive approach to group recommendations
  in the travel and tourism domain,'' in \emph{Adjunct Publication of the 25th
  conference on user Modeling, Adaptation and Personalization}, 2017, pp.
  11--16.

\bibitem{MG-UMUAI2006}
J.~Masthoff and A.~Gatt, ``In pursuit of satisfaction and the prevention of
  embarrassment: Affective state in group recommender systems,'' \emph{User
  Modeling and User-Adapted Interaction}, vol.~16, no. 3-4, pp. 281--319, 2006.

\bibitem{QSTIST2013}
L.~Quijano-Sanchez, J.~A. Recio-Garcia, B.~Diaz-Agudo, and G.~Jimenez-Diaz,
  ``Social factors in group recommender systems,'' \emph{ACM Transactions on
  Intelligent Systems and Technology}, vol.~4, no.~1, p.~8, 2013.

\bibitem{empathetic2014}
A.~Salehi-Abari and C.~Boutilier, ``Empathetic social choice on social
  networks,'' in \emph{Proceedings of The 13th International Conference on
  Autonomous Agents and Multiagent Systems}, ser. AAMAS'14, 2014, pp. 693--700.

\bibitem{delic2018use}
A.~Delic, J.~Masthoff, J.~Neidhardt, and H.~Werthner, ``How to use social
  relationships in group recommenders: Empirical evidence,'' in
  \emph{Proceedings of the 26th Conference on User Modeling, Adaptation and
  Personalization}, 2018, pp. 121--129.

\bibitem{delic2016observing}
A.~Delic, J.~Neidhardt, T.~N. Nguyen, F.~Ricci, L.~Rook, H.~Werthner, and
  M.~Zanker, ``Observing group decision making processes,'' in
  \emph{Proceedings of the 10th ACM Conference on Recommender Systems}, 2016,
  pp. 147--150.

\bibitem{hamilton2017inductive}
W.~Hamilton, Z.~Ying, and J.~Leskovec, ``Inductive representation learning on
  large graphs,'' in \emph{Proceeding of the Thirty-first Annual Conference on
  Neural Information Processing Systems (NIPS)}, 2017, pp. 1024--1034.

\bibitem{s_r+c_f+z_g+l_s+2012}
S.~Rendle, C.~Freudenthaler, Z.~Gantner, and L.~Schmidt-Thieme, ``{BPR}:
  Bayesian personalized ranking from implicit feedback,'' in \emph{Conference
  on Uncertainty in Artificial Intelligence (UAI)}, 2009, pp. 452--461.

\bibitem{Arrow1963SocialChoice}
K.~J. Arrow, \emph{Social Choice and Individual Values}.\hskip 1em plus 0.5em
  minus 0.4em\relax Yale University Press, 1951.

\bibitem{sen70Collective}
A.~K. Sen, \emph{Collective Choice and Social Welfare}.\hskip 1em plus 0.5em
  minus 0.4em\relax North-Holand, 1970.

\end{thebibliography}

\end{document}